\documentclass[10pt,journal,compsoc]{IEEEtran}
\ifCLASSOPTIONcompsoc
  \usepackage[nocompress]{cite}
\else
  \usepackage{cite}
\fi
\ifCLASSINFOpdf
\else
\fi

\usepackage{url}            % simple URL typesetting
\usepackage{booktabs}       % professional-quality tables
\usepackage{amsfonts}       % blackboard math symbols
\usepackage{nicefrac}       % compact symbols for 1/2, etc.
\usepackage{microtype}      % microtypography
\usepackage{epsfig}
\usepackage{graphicx}
\usepackage{amsmath}
\usepackage{amssymb}
\usepackage{mathrsfs}
\usepackage[ruled]{algorithm2e}
\usepackage[noend]{algpseudocode}
\usepackage{multirow}
\usepackage{subfigure}
\usepackage{rotating}
\usepackage{xcolor}
\usepackage{caption}
\usepackage{wrapfig}
\usepackage{lipsum}
 
\DeclareMathOperator*{\argmax}{arg\,max}

\newcommand{\R}{\mathbb{R}}

\begin{document}
\title{Out-of-Domain Generalization from a Single Source: An Uncertainty Quantification Approach}

\author{Xi Peng, Fengchun Qiao and Long Zhao     
\IEEEcompsocitemizethanks{
\IEEEcompsocthanksitem Xi Peng is with the Department of Computer and Information Sciences, University of Delaware. DE, USA. E-mail: xipeng@udel.edu.\protect\\
\IEEEcompsocthanksitem Fengchun Qiao is with the Department of Computer and Information Sciences, University of Delaware. DE, USA. E-mail: fengchun@udel.edu.\protect\\
\IEEEcompsocthanksitem Long Zhao is with the Department of Computer Science, Rutgers University, NJ, USA. E-mail: lz311@cs.rutgers.edu.}}


\markboth{Journal of \LaTeX\ Class Files,~Vol.~14, No.~8, August~2015}%
{Shell \MakeLowercase{\textit{et al.}}: Bare Advanced Demo of IEEEtran.cls for IEEE Computer Society Journals}



\IEEEtitleabstractindextext{%
\begin{abstract}

We are concerned with a worst-case scenario in model generalization, in the sense that a model aims to perform well on many unseen domains while there is only one single domain available for training. We propose Meta-Learning based Adversarial Domain Augmentation to solve this Out-of-Domain generalization problem. The key idea is to leverage adversarial training to create ``fictitious'' yet ``challenging'' populations, from which a model can learn to generalize with theoretical guarantees. To facilitate fast and desirable domain augmentation, we cast the model training in a meta-learning scheme and use a Wasserstein Auto-Encoder to relax the widely used worst-case constraint. We further improve our method by integrating uncertainty quantification for efficient domain generalization. Extensive experiments on multiple benchmark datasets indicate its superior performance in tackling single domain generalization.

\end{abstract}
}

\maketitle

% is not selected <OR> if conference mode is selected - because compsoc
% conference papers position the abstract like regular (non-compsoc)
\IEEEdisplaynontitleabstractindextext

% \fi
\IEEEpeerreviewmaketitle

\ifCLASSOPTIONcompsoc
\IEEEraisesectionheading{\section{Introduction}
\label{sec:introduction}}
\else
\section{Introduction}
\label{sec:introduction}
\fi

Recent years have witnessed rapid deployment of machine learning models for broad applications~\cite{lecun2015deep,shrivastava2017learning,konstantinos2018using,zhao2019semantic}. A key assumption underlying the remarkable success is that the training and test data usually follow similar statistics. Otherwise, even strong models ({\it e.g.,} deep neural networks) may break down on unseen or Out-of-Domain (OOD) test samples~\cite{balaji2018metareg}. Incorporating data from multiple training domains somehow alleviates this issue~\cite{li2017deeper}, but this may not always be applicable due to limited data acquiring budgets or privacy issues. An interesting yet seldom investigated problem then arises: Can a model generalize from one source domain to many unseen target domains? In other words, how to maximize the model generalization when there is only a single domain available for training?

\begin{figure}[!t]
\centering
\includegraphics[width=0.46\textwidth]{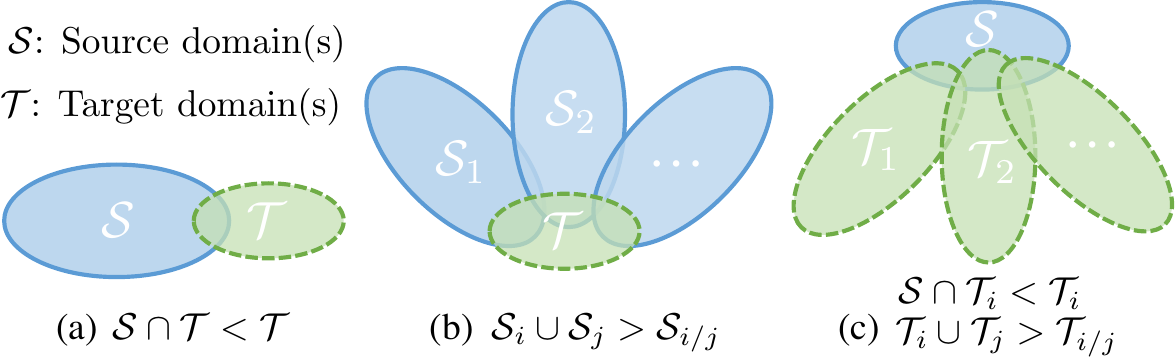}
\caption{The domain discrepancy: (a) domain adaptation, (b) domain generalization, and (c) single domain generalization.}
\label{fig:problem_illustration}
\end{figure}

The discrepancy between source and target domains, also known as domain or covariate variant~\cite{storkey2006mixture}, has been intensively studied in {\it domain adaptation}~\cite{motiian2017few,murez2018image,xu2019dsne,liu2019transferable} and {\it domain generalization}~\cite{muandet2013domain,ghifary2015domain,li2018learning,carlucci2019jigasaw}. Despite of their various success in tackling ordinary domain discrepancy issues, we argue that existing methods can hardly succeed in the aforementioned single domain generalization problem. As illustrated in Fig.~\ref{fig:problem_illustration}, the former usually expects the availability of target domain data (either labeled or unlabeled); While the latter, on the other hand, always assumes multiple (rather than one) domains are available for training. This fact emphasizes the necessity to develop a new learning paradigm for {\it single domain generalization}.

In this paper, we propose {\it adversarial domain augmentation} (Sec.~\ref{sec:ada}) to solve this challenging task. Inspired by the recent success in adversarial training~\cite{peng2018jointly,tang2019adaptive,szegedy2014intriguing,ratner2017learning,liu2019transferable}, we cast the single domain generalization problem in a worst-case formulation~\cite{sinha2017certifying,lee2018minimax}. The goal is to use single source domain to generate ``fictitious'' yet ``challenging'' populations, from which a model can learn to generalize with theoretical guarantees (Sec.~\ref{sec:theory}).

However, technical barriers exist when applying adversarial training for domain augmentation. On the one hand, it is hard to create ``fictitious'' domains that are largely different from the source, due to the contradiction of semantic consistency constraint~\cite{goodfellow2014adv} in worst-case formulation. On the other hand, we expect to explore many ``fictitious'' domains to guarantee sufficient coverage, which may result in significant computational overhead. To circumvent these barriers, we propose to {\it relax the worst-case constraint} (Sec.~\ref{sec:3.2}) via a Wasserstein Auto-Encoder (WAE)~\cite{tolstikhin2018wasserstein} to encourage large domain transportation in the input space. Moreover, rather than learning a series of ensemble models~\cite{volpi2018generalizing}, we organize adversarial domain augmentation via {\it meta-learning}~\cite{finn2017model} (Sec.~\ref{sec:meta}), yielding a highly efficient model with improved single domain generalization.

The capability of model generalization is usually measured in terms of accuracy. However, uncertainty quantification plays a vital role in mission-critical applications~\cite{finn2018probabilistic}. 
For instance, when deploying self-driving cars in unknown environments, it is crucial to be aware of the predictive uncertainty in risk assessment. Existing works~\cite{muandet2013domain,ghifary2015domain,li2018learning,carlucci2019jigasaw,dou2019domain} usually overlook the potential risk of leveraging augmented data in tackling out-of-domain generalization, raising serious safety and security concerns. To this end, we improve our method by integrating uncertainty quantification for broad and safe domain generalization.
To summarize, our contribution is multi-fold:
\begin{itemize}
    \item  The primary contribution of this work is a meta-learning based scheme that enables single domain generalization, an important yet seldom studied problem. We achieve the goal by proposing adversarial domain augmentation, while at the same time, relaxing the widely used worst-case constraint.
    \item To the best of our knowledge, we are the first to quantify the generalization uncertainty from a single source. We leverage the uncertainty assessment to increase the capacity in both input and label spaces.
    \item Extensive experiments indicate that our method marginally outperforms state of the art in single domain generalization of benchmark datasets including {\it Digits}, {\it CIFAR-10-C}~\cite{hendrycks2019benchmarking}, and {\it SYTHIA}~\cite{ros2016synthia}.
\end{itemize}

This work features substantial extensions to our conference version~\cite{qiao2020learning}.
We propose to jointly perturb both latent features and ground-truth labels to encourage significant domain augmentations, integrating uncertainty quantification for more reliable
out-of-domain generalization.

\section{Related Work}

In this section, we first review recent progress on domain adaptation and domain generalization, especially in the area of object recognition. Then, we briefly discuss previous works on adversarial training, meta-learning and uncertainty assessment, which are closely related to our method.

{\bf Domain adaptation.}
Domain discrepancy brought by domain or covariance shifts~\cite{storkey2006mixture} severely degrades the model performance on cross-domain recognition. 
The models trained using Empirical Risk Minimization~\cite{koltchinskii2011oracle} usually perform poorly on unseen domains. 
To reduce the discrepancy across domains, a series of methods are proposed for unsupervised~\cite{murez2018image,shu2018dirt,french2017self,russo2018source,sankaranarayanan2018generate} or supervised domain adaptation~\cite{motiian2017unified,xu2019dsne}. In unsupervised domain adaptation, works in this field mostly attempt to match the distribution of the target domain to that of the source domain by minimizing the maximal mean discrepancy~\cite{russo2018source} or a domain classifier network~\cite{liu2019transferable}.
In supervised domain adaptation, some recent work focused on few-shot domain adaptation~\cite{motiian2017few} where only a few labeled samples from target domain are involved in training. The source and target domains are strongly coupled in both scenarios, limiting their generalization ability to other target domains.
In this work, we expect the model to generalize well on many unseen target domains.

 {\bf Domain generalization.}
Domain generalization~\cite{ghifary2015domain,li2017deeper,grubinger2017multi,shankar2018generalizing,carlucci2019jigasaw,dou2019domain} is another line of research, which has been intensively studied in recent years. 
Different from domain adaptation, domain generalization aims to learn from multiple source domains without any access to target domains.
Most previous methods either tried to learn a domain-invariant space to align domains~\cite{muandet2013domain,ghifary2015domain,grubinger2017multi,li2017deeper,zhao2020knowledge} or aggregate domain-specific modules~\cite{mancini2018robust,mancini2018best}. 
A few studies~\cite{carlucci2019jigasaw,shankar2018generalizing,volpi2018generalizing,qiao2020learning,zhao2020maximum} proposed data augmentation strategies to generate additional training data to enhance the generalization capabiltiy over unseen domains.
JiGen~\cite{carlucci2019jigasaw} proposed to generate jigsaw puzzles from source domains and leverage them as self-supervised signals. CrossGrad~\cite{shankar2018generalizing} augments input instances with domain-guided perturbations through an auxiliary domain classifier.
GUD~\cite{volpi2018generalizing} proposed adversarial data augmentation to solve single domain generalization, and learned an ensemble model for stable training. 
Compared to GUD~\cite{volpi2018generalizing}, this work aims at creating large domain transportation for ``fictitious'' domains and devising a more efficient meta-learning scheme within a single unified model.
Several methods~\cite{madry2017pgd,wang2019learning,hendrycks2019using} proposed to leverage adversarial training~\cite{goodfellow2014adv} to learn robust models, which can also be applied in single domain generalization. PAR~\cite{wang2019learning} proposed to learn robust global representations by penalizing the predictive power of local representations. Hendrycks \textit{et al.}~\cite{hendrycks2019using} applied self-supervised learning to improve the model robustness.

{\bf Adversarial training.} 
Adversarial training was originally proposed by~\cite{szegedy2014intriguing}. They discovered that deep neural networks, including state-of-the-art models, are particularly vulnerable to minor adversarial perturbations.
Goodfellow \textit{et al.}~\cite{goodfellow2014adv} proposed Fast Gradient Sign Method (FGSM) which takes the sign of a gradient obtained from the classifier. FGSM made the model robust against adversarial samples and improved generalization performance.
Madry \textit{et al.}~\cite{madry2017pgd} illustrated that adversarial samples generated through 
projected gradient descent can provide robustness guarantees. 
Miyato \textit{et al.}~\cite{miyato2018virtual} proposed virtual adversarial training to smooth the output distributions as a regularization of models.
Sinha \textit{et al.}~\cite{sinha2017certifying} proposed principled adversarial training with robustness guarantees through distributionally robust optimization. 
More recently, Stutz \textit{et al.}~\cite{stutz2019disentangling} illustrated that on-manifold adversarial samples can improve generalization. Therefore, models with both robustness and strong generalization capability can be achieved at the same time. 
In our work, we leverage adversarial training to create augmentations with large domain transportation.

{\bf Meta-learning.}
Meta-learning~\cite{schmidhuber1987evolutionary,thrun2012learning} is a long standing topic on learning models to generalize over a distribution of tasks. 
It has been widely used in optimization of deep neural networks~\cite{andrychowicz2016learning,lilearning} and few-shot classification~\cite{koch2015siamese,vinyals2016matching,snell2017prototypical}.
Recently, Finn \textit{et al.}~\cite{finn2017model} proposed a Model-Agnostic Meta-Learning (MAML) procedure for few-shot learning and reinforcement learning.
The objective of MAML is to find a good initialization which can be fast adapted to new tasks within few gradient steps. 
In this paper, we propose a modified MAML to make the model generalize over the distribution of domain augmentation.
Several approaches~\cite{li2018learning,balaji2018metareg,dou2019domain} have been proposed to learn domain generalization in a meta-learning framework. 
Li \textit{et al.}~\cite{li2018learning} firstly applied MAML in domain generalization by adopting an episodic training paradigm. 
Balaji \textit{et al.}~\cite{balaji2018metareg} proposed to meta-learn a regularization function to train networks which can be easily generalized to different domains. Dou \textit{et al.}~\cite{dou2019domain} incorporated global and local constraints for learning semantic feature spaces in a meta-learning framework.
However, these methods cannot be directly applied for single source generalization since there is only one distribution available during training. 

{\bf Uncertainty quantification.}
MC-dropout~\cite{gal2016dropout} leverages the standard dropout layer~\cite{srivastava2014dropout} to estimate the model uncertainty during inference time. 
Bayesian neural networks~\cite{hinton1993keeping,graves2011practical,blundell2015weight} provide a natural way to integrate uncertainty into weights of deep networks.
Several Bayesian meta-learning frameworks~\cite{grant2018recasting,finn2018probabilistic,yoon2018bayesian,lee2019learning} have been proposed to model the uncertainty of few-shot tasks. 
Grant \textit{et al.}~\cite{grant2018recasting} proposed the first Bayesian variant of MAML~\cite{finn2017model} using the Laplace approximation. 
Finn \textit{et al.}~\cite{finn2018probabilistic} approximated MAP inference of the task-specific weights while maintain uncertainty only in the global weights.
Lee \textit{et al.}~\cite{lee2019meta} proposed meta-dropout which generates learnable perturbations to regularize few-shot learning models.
In this paper, instead of estimating the uncertainty of tasks, we propose to quantify the uncertainty of domains and leverage it to guide model generalization.

\section{Single Domain Generalization}
We aim at solving the problem of single domain generalization: 
A model is trained on only one source domain $\mathcal{S}$ but is expected to generalize over a {\it unknown} domain distribution $\{\mathcal{T}_1,\mathcal{T}_2,\cdots\} \sim p(\mathcal{T})$. This problem is more challenging than {\it domain adaptation} (assume $p(\mathcal{T})$ is given) and {\it domain generalization} (assume multiple source domains $\{\mathcal{S}_1,\mathcal{S}_2,\cdots\}$ are available). A promising solution of this challenging problem, inspired by many recent achievements~\cite{ratner2017learning,volpi2018generalizing,liu2019transferable}, is to leverage adversarial training~\cite{goodfellow2014adv,szegedy2014intriguing}. The key idea is to learn a robust model that is resistant to out-of-distribution perturbations.
More specifically, we can learn the model by solving a worst-case problem~\cite{sinha2017certifying}: 
\begin{equation}
\underset{\theta}{\operatorname{min}} \sup _{\mathcal{T}: D\left(\mathcal{S},\mathcal{T}\right) \leq \rho} \mathbb{E}[\mathcal{L}_{\mathrm{task}}(\theta ;\mathcal{T})],
\label{eq:worst1}
\end{equation}
where $D$ is a similarity metric to measure the domain distance and $\rho$ denotes the largest domain discrepancy between $\mathcal{S}$ and $\mathcal{T}$. $\theta$ are model parameters that are optimized according to a task-specific objective function $\mathcal{L}_{\mathrm{task}}$. Here, we focus on classification problems using cross-entropy loss:
\begin{equation}\label{eq:ce}
\mathcal{L}_{\mathrm{task}}(\mathbf{y}, \hat{\mathbf{y}}) = -\sum_i y_i \log (\hat{y}_{i}),
\end{equation}
where $\hat{\mathbf{y}}$ is \textit{softmax} output of the model; $\mathbf{y}$ is the one-hot vector representing the ground truth class; $y_i$ and $\hat{y}_{i}$ represent the $i$-th dimension of $\mathbf{y}$ and $\hat{\mathbf{y}}$, respectively.

\begin{figure}[t]
\centering
\includegraphics[width=1.0\linewidth]{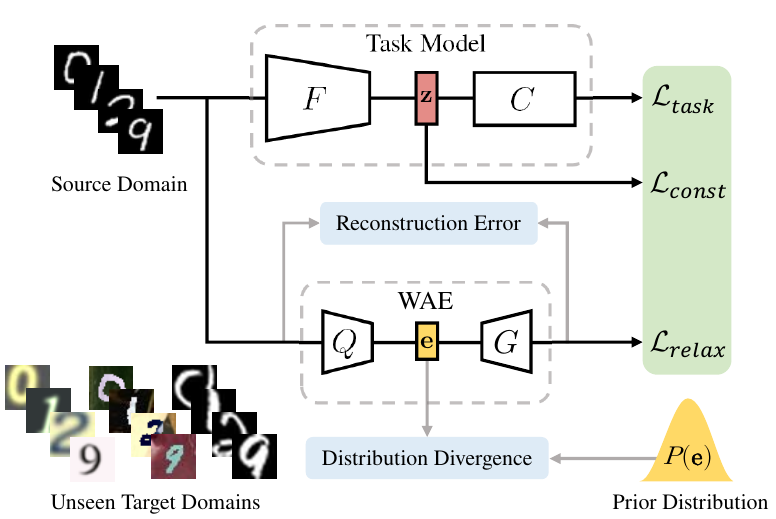}
\caption{Overview of adversarial domain augmentation. WAE is used to relax the semantic constraint and encourages large domain transportation.}
\label{fig_model}
\end{figure}

Following the worst-case formulation~\eqref{eq:worst1}, we propose a new method, \textit{Meta-Learning based Adversarial Domain Augmentation}, for single domain generalization. Fig.~\ref{fig_model} presents an overview of our approach. We create ``fictitious'' yet ``challenging'' domains by leverage adversarial training to augment the source domain in Sec.~\ref{sec:ada}. The task model learns from the domain augmentations with the assistance of a Wasserstein Auto-Encoder (WAE), which relaxes the worst-case constraint in Sec.~\ref{sec:3.2}. We organize the joint training of task model and WAE, as well as the domain augmentation procedure, in a meta-learning framework as described in Sec.~\ref{sec:meta}. 

\subsection{Adversarial Domain Augmentation} \label{sec:ada}
Our goal is to create multiple augmented domains from the source domain. Augmented domains are required to be distributionally different from the source domain so as to mimic unseen domains. In addition, to avoid divergence of augmented domains, the worst-case guarantee defined in Eq.~\eqref{eq:worst1} should also be satisfied.

To achieve this goal, we propose Adversarial Domain Augmentation. Our model consists of a task model and a WAE shown in Fig.~\ref{fig_model}. In Fig.~\ref{fig_model}, the task model consists of a feature extractor $F: \mathcal{X} \to \mathcal{Z}$ mapping images from input space to embedding space, and a classifier $C: \mathcal{Z} \to \mathcal{Y}$ used to predict labels from embedding space. Let $\mathbf{z}$ denote the latent representation of $\mathbf{x}$ which is obtained by $\mathbf{z} = F(\mathbf{x})$. The overall loss function is formulated as follows: 
\begin{equation}\label{eq:uada}
\mathcal{L}_\text{ADA} = \underbrace{\mathcal{L}_{\mathrm{task}}(\theta;\mathbf{x})}_{\mathrm{Classification}}  - \alpha \underbrace{\mathcal{L}_{\mathrm{const}}(\theta;\mathbf{z})}_{\mathrm{Constraint}}+\beta\underbrace{ \mathcal{L}_{\mathrm{relax}}(\psi;\mathbf{x})}_{\mathrm{Relaxation}}, 
\end{equation}
where $\mathcal{L}_{\mathrm{task}}$ is the classification loss defined in Eq.~\eqref{eq:ce}, $\mathcal{L}_{\mathrm{const}}$ is the worst-case guarantee defined in Eq.~\eqref{eq:worst1}, and $\mathcal{L}_{\mathrm{relax}}$ guarantees large domain transportation defined in Eq.~\eqref{eq:relax}. $\psi$ are parameters of the WAE.
$\alpha$ and $\beta$ are two hyper-parameter to balance $\mathcal{L}_{\mathrm{const}}$ and $\mathcal{L}_{\mathrm{relax}}$. 

Given the objective function $\mathcal{L}_\text{ADA}$, we employ an iterative way to generate the adversarial samples $\mathbf{x}^+$ in the augmented domain $\mathcal{S}^+$:
\begin{equation}
\mathbf{x}^+_{t+1} \gets \mathbf{x}^+_{t} +  \gamma\nabla_{\mathbf{x}^+_{t}}\mathcal{L}_\text{ADA}(\theta,\psi;\mathbf{x}^+_{t},\mathbf{z}^+_{t}),
\label{eq:ascent}
\end{equation}
where $\gamma$ is the learning rate of gradient ascent. A small number of iterations are required to produce sufficient perturbations and create desirable adversarial samples.

$\mathcal{L}_{\mathrm{const}}$ imposes semantic consistency constraint to adversarial samples so that $\mathcal{S}^+$ satisfies $D\left(\mathcal{S},\mathcal{S}^+\right)\leq \rho$.
More specifically, we follow~\cite{volpi2018generalizing} to measure the Wasserstein distance between $\mathcal{S}^+$ and $\mathcal{S}$ in the embedding space:
\begin{equation}\label{eq:constraint}
\mathcal{L}_{\mathrm{const}}= \frac{1}{2}\Vert \mathbf{z}-\mathbf{z}^+\Vert_{2}^{2}+\infty \cdot \mathbf{1}\left\{\mathbf{y} \neq \mathbf{y}^+\right\},
\end{equation}
where $\mathbf{1}\{\cdot\}$ is the 0-1 indicator function and $\mathcal{L}_{\mathrm{const}}$ will be $\infty$ if the class label of $\mathbf{x}^+$ is different from $\mathbf{x}$. 
We assume $y^+ == y$ is always true given a small enough step size, which simplifies implementation without compromising performance.
Intuitively, $\mathcal{L}_{\mathrm{const}}$ controls the ability of generalization outside the source domain measured by Wasserstein distance~\cite{villani2003topics}.
In the conventional setting of adversarial training, the worst-case problem is handled by only $\mathcal{L}_{\mathrm{task}}$ and $\mathcal{L}_{\mathrm{const}}$.
However, $\mathcal{L}_{\mathrm{const}}$ yields limited domain transportation since it severely constrains the semantic distance between the samples and their perturbations. 
Hence, $\mathcal{L}_{\mathrm{relax}}$ is proposed to relax the semantic consistency constraint and create large domain transportation. 
The implementation of $\mathcal{L}_{\mathrm{relax}}$ is discussed in Sec.~\ref{sec:3.2}. 

\subsection{Relaxation of Wasserstein Distance Constraint}\label{sec:3.2}
\label{sec_wdis}

\begin{figure}[t]
\centering
\includegraphics[width=1.0\linewidth]{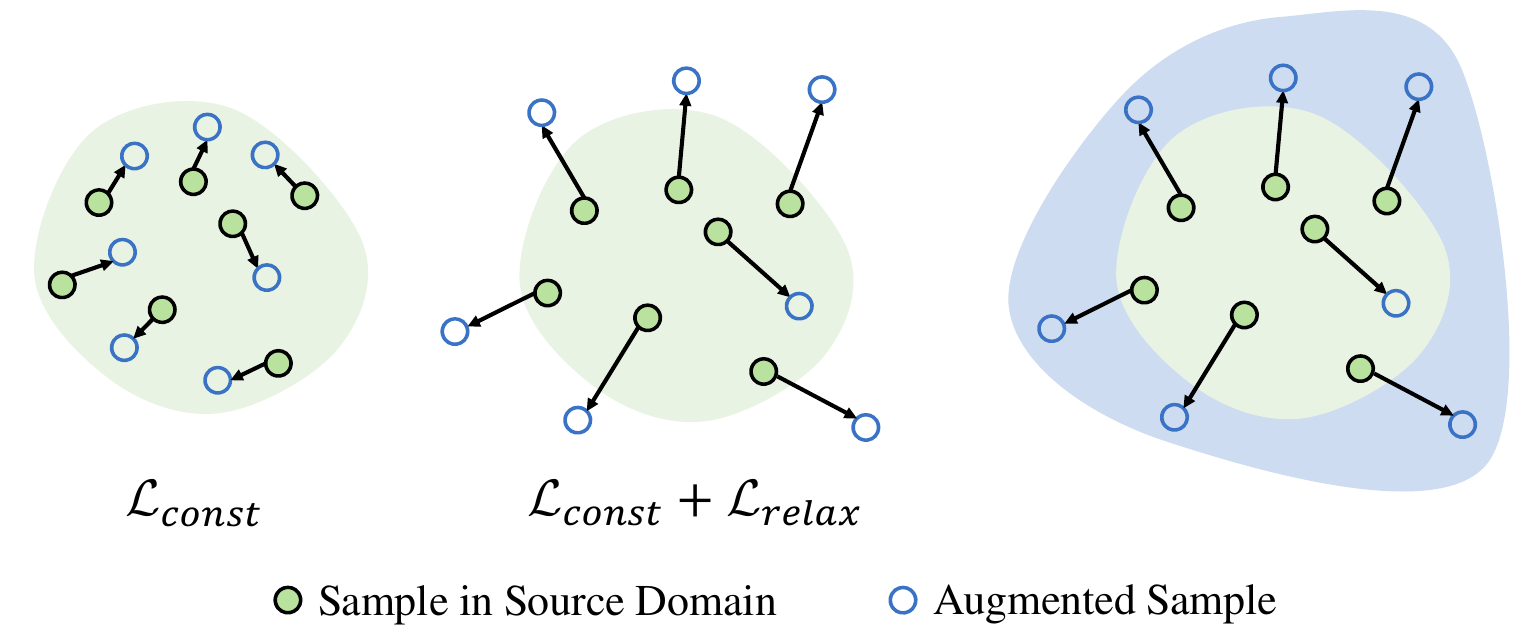}
\caption{Motivation of $\mathcal{L}_{\mathrm{relax}}$. {\bf Left:} The augmented samples may be close to the source domain if applying $\mathcal{L}_{\mathrm{const}}$. {\bf Middle:} We expect to create out-of-domain augmentations by incorporating $\mathcal{L}_{\mathrm{relax}}$. {\bf Right:} This would yield an enlarged training domain.}
\label{fig_ada}
\end{figure}

Intuitively, we expect the augmented domains $\mathcal{S}^+$ are largely different from the source domain $\mathcal{S}$.
In other words, we want to maximize the domain discrepancy between $\mathcal{S}^+$ and $\mathcal{S}$.
However, the semantic consistency constraint $\mathcal{L}_{\mathrm{const}}$ would severely limits the domain transportation from $\mathcal{S}$ to $\mathcal{S}^+$, posing new challenges to generate desirable $\mathcal{S}^+$. To address this issue, we propose $\mathcal{L}_{\mathrm{relax}}$ to encourage out-of-domain augmentations.
We illustrate the idea in Fig.~\ref{fig_ada}.

Specifically, we employ Wasserstein Auto-Encoders (WAEs)~\cite{tolstikhin2018wasserstein} to implement $\mathcal{L}_{\mathrm{relax}}$. Let $V$ denote the WAE parameterized by $\psi$. $V$ consists of an encoder $Q(\mathbf{e}|\mathbf{x})$ and a decoder $G(\mathbf{x}|\mathbf{e})$ where $\mathbf{x}$ and $\mathbf{e}$ denote inputs and bottleneck embedding, respectively. Additionally, we use a distance metric $\mathcal{D}_{\mathbf{e}}$ to measure the divergence between $Q(\mathbf{x})$ and a prior distribution $P(\mathbf{e})$, which can be implemented as either \textit{Maximum Mean Discrepancy} (MMD) or GANs \cite{goodfellow2014generative}. We can learn $V$ by optimizing:
\begin{equation}
\min _{\psi} [ \Vert G(Q(\mathbf{x}))- \mathbf{x} \Vert^2 + \lambda \mathcal{D}_{\mathbf{e}} (Q(\mathbf{x}),P(\mathbf{e}))],
\label{eq:wae}
\end{equation}
where $\lambda$ is a hyper-parameter. After pre-training $V$ on the source domain $S$ offline, we keep it frozen and maximize the reconstruction error $\mathcal{L}_{\mathrm{relax}}$ for domain augmentation:
\begin{equation}
\mathcal{L}_{\mathrm{relax}}= \Vert \mathbf{x}^+- V(\mathbf{x}^+) \Vert^2.
\label{eq:relax}
\end{equation}

Different from Vanilla or Variation Auto-Encoders \cite{kingma2013auto}, WAEs employ the Wasserstein metric to measure the distribution distance between the input and reconstruction. Hence, the pre-trained $V$ can better capture the distribution of the source domain and maximizing $\mathcal{L}_{\mathrm{relax}}$ creates large domain transportation. 

In this work, $V$ acts as a \textit{one-class discriminator} to distinguish whether the augmentation is outside the source domain, which is significantly different from the traditional discriminator of GANs \cite{goodfellow2014generative}. And it is also different from the domain classifier widely used in domain adaptation \cite{liu2019transferable}, since there is only one source domain available. As a result, $\mathcal{L}_{\mathrm{relax}}$ together with $\mathcal{L}_{\mathrm{const}}$ are used to ``push away'' $\mathcal{S}^+$ in input space and ``pull back'' $\mathcal{S}^+$ in the embedding space simultaneously. In Sec.~\ref{sec:theory}, we show that $\mathcal{L}_{\mathrm{relax}}$ and $\mathcal{L}_{\mathrm{const}}$ are derivations of two Wasserstein distance metrics defined in the input space and embedding space, respectively.

\subsection{Meta-Learning Single Domain Generalization}\label{sec:meta}

To efficiently organize the model training on the source domain $S$ and augmented domains $\mathcal{S}^+$, we leverage a meta-learning scheme to train a single model. To mimic real domain-shifts between the source domain $S$ and target domain $T$, at each learning iteration, we perform meta-train on the source domain $\mathcal{S}$ and meta-test on all augmented domains $\mathcal{S}^+$. Hence, after many iterations, the model is expected to achieve good generalization on the final target domain $\mathcal{T}$ during evaluation.

Formally, the proposed Meta-Learning based Adversarial Domain Augmentation approach consists of three parts in each iteration during the training procedure: meta-train, meta-test and meta-update. In meta-train, $\mathcal{L}_{\mathrm{task}}$ is computed on samples from the source domain $\mathcal{S}$, and the model parameters $\theta$ is updated via one or more gradient steps with a learning rate of $\eta$:
\begin{equation}\label{eq:meta-train}
\hat{\theta} \gets \theta - \eta \nabla_{\theta} \mathcal{L}_{\mathrm{task}}(\theta; \mathcal{S}).
\end{equation}
Then we compute $\mathcal{L}_{\mathrm{task}}(\hat{\theta}; \mathcal{S}^+_k)$ on each augmented domain $\mathcal{S}^+_k$ in meta-test. At last, in meta-update, we update $\theta$ by the gradients calculated from a combined loss where  meta-train and meta-test are optimised simultaneously:
\begin{equation}\label{eq:meta-update}
\theta \gets \theta - \eta\nabla_{\theta} [\mathcal{L}_{\mathrm{task}}(\theta; \mathcal{S})+ \sum^{K}_{k=1}\mathcal{L}_{\mathrm{task}}(\hat{\theta}; \mathcal{S}^+_k)],
\end{equation}
where $K$ is the number of augmented domains. 
Note that in addition to the augmented domains $\mathcal{S}^+$, we also minimize the loss on the source domain $\mathcal{S}$ to avoid performance degradation when $\mathcal{S}^+$ are far away from $\mathcal{S}$.

The entire training pipeline of {\it Single Domain Generalization} (SDG) is summarized in Alg.~\ref{alg:overrall}.
In contrast to prior work~\cite{volpi2018generalizing} that learns a series of ensemble models, our method achieves a single model for efficiency. More importantly, the meta-learning scheme prepares the learned model for fast adaptation: One or a small number of gradient steps will produce improved behavior on a new target domain. This enables our method for \textit{few-shot domain adaptation}. Please refer to Sec~\ref{sec:few} for more details.

\section{Uncertain Single Domain Generalization}\label{sec:uncertain}

\begin{figure}
    \centering
    \includegraphics[width=.6\linewidth]{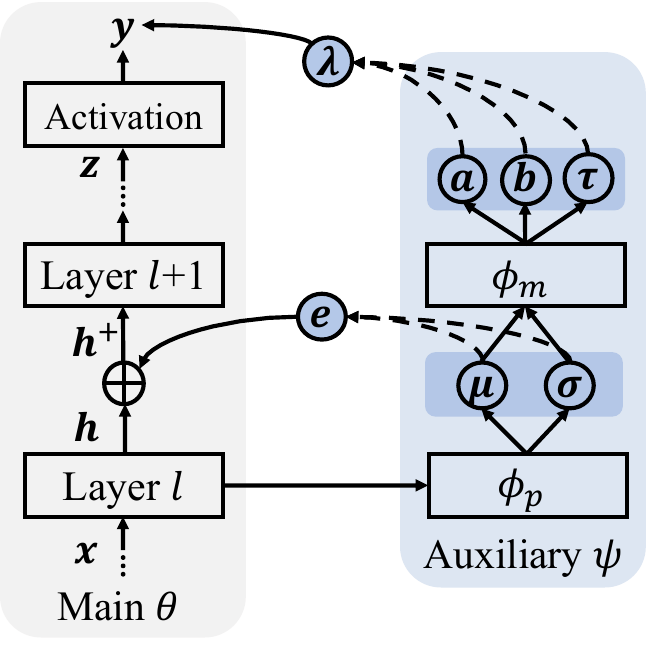}
    \caption{Overview of uncertainty-guided domain augmentation. The auxiliary model $\phi$ is used to create uncertainty-guided augmentations in both feature and label spaces.}\label{fig:models}
\end{figure}

Although our method improves the generalization capability in terms of accuracy, it ignores the the potential risk of leveraging augmented data in tackling out-of-domain generalization. 
To this end, we propose {\it Uncertain Single Domain Generalization} (Uncertain SDG) to further improve it by integrating uncertainty quantification for efficient domain generalization. To achieve this, we introduce the auxiliary model $\psi=\{\phi_p,\phi_m\}$ to explicitly model the uncertainty with respect to $\theta$ and gradually increase the uncertainty to create more challenging $\mathcal{S}^+$ by increasing the capacity in both input and label spaces. In input space, we introduce $\phi_p$ to create feature augmentations $\mathbf{h}^{+}$ via adding perturbation $\mathbf{e}$ sampled from $\mathcal{N}(\boldsymbol{\mu},\boldsymbol{\sigma})$. In label space, we integrate the same uncertainty encoded in  $(\boldsymbol{\mu},\boldsymbol{\sigma})$ into $\phi_m$ and propose learnable mixup to generate $\mathbf{y}^{+}$ (together with $\mathbf{h}^{+}$) through three variables $(a,b,\tau)$, yielding consistent augmentation in both input and output spaces.

\noindent{\bf Input Augmentation.}
The goal is to assess the uncertainty of $\mathcal{S}^+$ and use it to provide consistent augmentation in both input and output spaces.
Towards this goal, instead of directly augmenting raw data, we introduce a light auxiliary network $\phi_p$ to create feature augmentation $\mathbf{h}^+$  with large domain transportation through increasing the uncertainty with respect to $\theta$.
We propose to learn layer-wise feature perturbations $\mathbf{e}$ that transport latent features $\mathbf{h} \rightarrow \mathbf{h}^+$ for efficient domain augmentation $\mathcal{S} \rightarrow \mathcal{S}^+$. Instead of a direct generation $\mathbf{e}=f_{\phi_p}(\mathbf{x},\mathbf{h})$ widely used in previous work~\cite{volpi2018generalizing,qiao2020learning}, we assume $\mathbf{e}$ follows a multivariate Gaussian distribution $\mathcal{N}(\boldsymbol{\mu},\boldsymbol{\sigma})$, which can be used to easily access the uncertainty when deploying the model on unseen domains. 
More specifically, the Gaussian parameters are learnable via variational inference $(\boldsymbol{\mu},\boldsymbol{\sigma})=f_{\phi_p}(\mathcal{S},\theta)$, such that:
\begin{equation}
\mathbf{h}^+ \leftarrow \mathbf{h} + \text{Softplus}(\mathbf{e})
\text{, where }
\mathbf{e} \sim \mathcal{N}(\boldsymbol{\mu},\boldsymbol{\sigma}),
\label{eq:pertubation}
\end{equation}
where $\text{Softplus}(\cdot)$ is applied to stabilize the training.

The proposed uncertainty-guided input augmentation is capable of relaxing the worst-case constraint in Eq.~\eqref{eq:uada} and enlarging the domain transportation, which is more
efficient than SDG that has to train an extra Wasserstein Auto-Encoder~\cite{tolstikhin2018wasserstein} to achieve this goal. In Sec.~\ref{sec:ablation}, we empirically show that $\{\mathbf{h}^+_1, \mathbf{h}^+_2,\cdots\}$ gradually enlarge the transportation by increasing the uncertainty of augmentations, enabling the model to learn from ``easy'' to ``hard'' domains in a curriculum learning scheme. In Sec.~\ref{sec:sota}, we empirically demonstrate that Uncertain SDG outperforms SDG marginally in terms
of memory, speed and accuracy.

\noindent{\bf Label Augmentation.}
Feature perturbations do not only augment the input but also yield label uncertainty. 
To explicitly model the label uncertainty, we leverage the input uncertainty, encoded in  $(\boldsymbol{\mu},\boldsymbol{\sigma})$, 
to infer the label uncertainty encoded in $(a,b,\tau)$ through $\phi_m$ as shown in Fig.~\ref{fig:models}.
Inspired by {\it mixup} which performs convex interpolations of pairs of examples ($\mathbf{x}_i,\mathbf{x}_j$) and their labels ($\mathbf{y}_i,\mathbf{y}_j$), we improve {\it mixup} by casting it in a learnable framework specially tailored for single source generalization.
First, instead of mixing up pairs of examples, we mix up $\mathcal{S}$ and $\mathcal{S}^+$ to achieve in-between domain interpolations. Second, we leverage the uncertainty encoded in $(\boldsymbol{\mu},\boldsymbol{\sigma})$ to predict learnable parameters $(a,b)$, which controls the direction and strength of domain interpolations: 
\begin{equation}\label{eq:mixup}
\mathbf{h}^+ =\lambda \mathbf{h} + (1-\lambda) \mathbf{h}^+, \quad \mathbf{y}^+ =\lambda \mathbf{y}+(1-\lambda) \tilde{\mathbf{y}},
\end{equation}
where $\lambda \sim \operatorname{Beta}(a, b)$ and $\tilde{\mathbf{y}}$ denotes a {\it label-smoothing}~\cite{szegedy2016rethinking} version of $\mathbf{y}$. More specifically, we perform {\it label smoothing} by a chance of $\tau$, such that we assign $\rho \in(0,1)$ to the true category and equally distribute $\frac{1-\rho}{c-1}$ to the others, where $c$ counts categories. The Beta distribution $(a,b)$ and the lottery $\tau$ are jointly inferred by $(a,b,\tau)=f_{\phi_m}(\boldsymbol{\mu},\boldsymbol{\sigma})$ to integrate the uncertainty of domain augmentation. In Sec.~\ref{sec:ablation}, we empirically prove that the uncertainty encoded in $\boldsymbol{\mu}$ and $\boldsymbol{\sigma}$ can encourage more smoothing labels and significantly increase the capacity of label space.
The entire training pipeline is summarized in Alg.~\ref{alg:uncertain}.

\begin{algorithm}[t]
	\caption{Single Domain Generalization (SDG).}
	\LinesNumbered
	\label{alg:overrall}
	\KwIn{Source domain $\mathcal{S}$; Pre-train WAE $V$ on $\mathcal{S}$; \# of augmented domains $K$}
	\KwOut{Learned model parameters $\theta$ }
	
	\For{$k=1,...,K$}{
		Generate $\mathcal{S}^+_{k}$ from $\mathcal{S} \cup \{\mathcal{S}^+_{i}\}^{k-1}_{i=1} $ using Eq.~\eqref{eq:ascent} \\
		Re-train $V$ with $\mathcal{S}^+_{k}$ \\
		\textbf{Meta-train}: Evaluate $\mathcal{L}_{\mathrm{task}}(\theta; \mathcal{S})$ w.r.t $\mathcal{S}$ \\
		Compute $\hat{\theta}$ on $\mathcal{S}$ using Eq.~\eqref{eq:meta-train} \\
		\For{$i=1,...,k$}{
			\textbf{Meta-test}: Evaluate $\mathcal{L}_{\mathrm{task}}(\hat{\theta}; \mathcal{S}^+_i)$ w.r.t $\mathcal{S}^+_i$ \\
		}
		\textbf{Meta-update}: Update $\theta$ using Eq.~\eqref{eq:meta-update}
	}
\end{algorithm}

\begin{algorithm}[t]
	\caption{Uncertain Single Domain Generalization (Uncertain SDG).}
	\LinesNumbered
	\label{alg:uncertain}
	\KwIn
	{Source domain $\mathcal{S}$, \# of augmented domains $K$.}
	\KwOut{Learned backbone $\theta$ and auxiliary $\psi$}
	 \While{not converged}{
	     \textbf{Meta-train}: Evaluate $\mathcal{L}_{\mathrm{task}}(\theta; \mathcal{S})$ w.r.t $\mathcal{S}$ \\
		Compute $\hat{\theta}$ on $\mathcal{S}$ using Eq.~\eqref{eq:meta-train} \\

	     \For{$k=1,...,K$}{
	     Sample feature perturbation $\mathbf{h}^+_k$ using Eq.~\eqref{eq:pertubation}
	     
	     Generate label mixup  $\mathbf{y}^+_k$  using Eq.~\eqref{eq:mixup}
	     
	     \textbf{Meta-test}: Evaluate $\mathcal{L}_{\mathrm{task}}(\hat{\theta};\mathcal{S}^+_{k}) $ w.r.t. $\mathcal{S}^+_{k}$
        }
        \textbf{Meta-update}: Update $\theta$ and $\psi$.
   }
\end{algorithm}

\begin{figure*}[t]
\begin{center}
\subfigure{
	\includegraphics[width=0.20\linewidth]{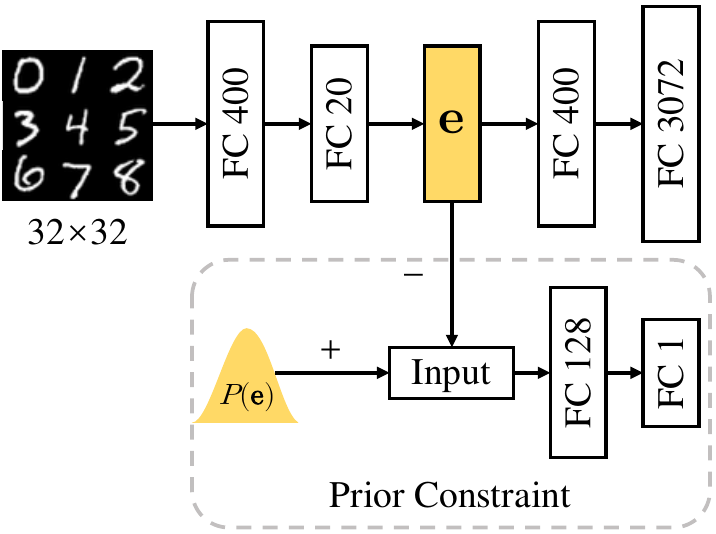}
}
\subfigure{
	\includegraphics[width=0.40\linewidth]{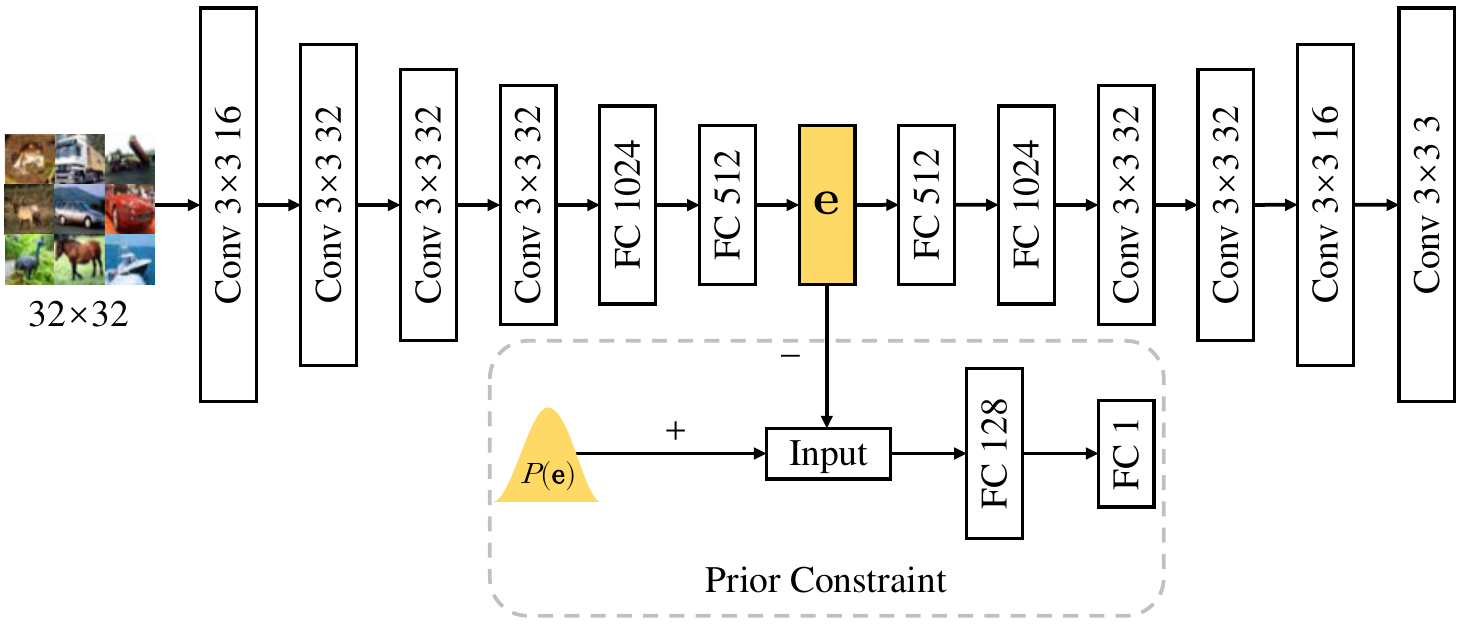}
}
\subfigure{
	\includegraphics[width=0.35\linewidth]{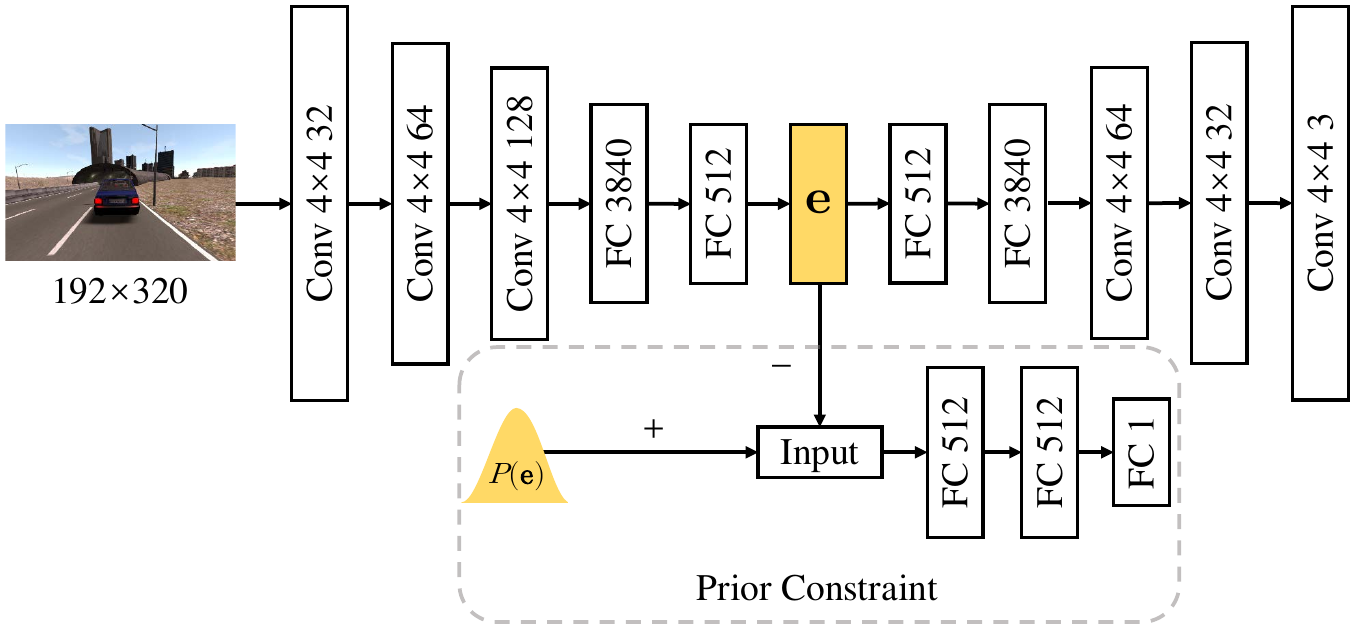}
}
\end{center}
\vspace{-1em}
\caption{Architectures of WAEs. \textbf{From left to right:} (a) WAE for {\it Digits }; (b) WAE for {\it CIFAR-10-C}~\cite{hendrycks2019benchmarking}; and (c) WAE for {\it SYTHIA}~\cite{ros2016synthia}.
Note that ``\textbf{+}'': positive samples for discriminator; ``\textbf{-}'': negative samples for discriminator. } 
\label{fig_waes}
\end{figure*}

\section{Theoretical Understanding} \label{sec:theory}
We provide a detailed theoretical analysis of the proposed Adversarial Domain Augmentation.
Specifically, we show that the overall loss function defined in Eq.~\eqref{eq:uada} is a direct derivation of a relaxed worst-case problem.

Let $c: \mathcal{Z} \times \mathcal{Z} \to \R_+ \cup \{\infty\}$ be the ``cost'' for an adversary to perturb $\mathbf{z}$ to $\mathbf{z}^+$ in the embedding space. Let $d: \mathcal{X} \times \mathcal{X} \to \R_+ \cup \{\infty\}$ be the ``cost'' for an adversary to perturb $\mathbf{x}$ to $\mathbf{x}^+$ in the input space. The Wasserstein distances between $\mathcal{S}$ and $\mathcal{S}^+$ can be formulated as:
$W_c(\mathcal{S}, \mathcal{S}^+):=\inf _{M_{\mathbf{z}} \in \Pi(\mathcal{S}, \mathcal{S}^+)} \mathbb{E}_{M_{\mathbf{z}}}\left[c\left(\mathbf{z}, \mathbf{z}^+\right)\right]$ and
$W_d(\mathcal{S}, \mathcal{S}^+):=\inf _{M_{\mathbf{x}} \in \Pi(\mathcal{S}, \mathcal{S}^+)} \mathbb{E}_{M_{\mathbf{x}}}\left[d\left(\mathbf{x}, \mathbf{x}^+\right)\right]$, where $M_{\mathbf{z}}$ and $M_{\mathbf{x}}$ are measures in the embedding and input space, respectively; $\Pi(\mathcal{S}, \mathcal{S}^+)$ is the joint distribution of $\mathcal{S}$ and $\mathcal{S}^+$.
Then, the relaxed worst-case problem can be formulated as:
\begin{equation}
\theta^\ast = \min_\theta \sup_{\mathcal{S}^+\in\mathcal{D}}\mathbb{E}[\mathcal{L}_{\mathrm{task}}(\theta;\mathcal{S}^+)],
\label{eq_problem}
\end{equation}
where $\mathcal{D} = \{\mathcal{S}^+: W_c(\mathcal{S}, \mathcal{S}^+) \le \rho, W_d(\mathcal{S}, \mathcal{S}^+) \ge \eta\}$. We note that $\mathcal{D}$ 
covers a robust region that is within $\rho$ distance of $\mathcal{S}$ in the embedding space and $\eta$ distance away from $\mathcal{S}$ in the input space under the Wasserstein distance measures $W_c$ and $W_d$, respectively.

\begin{figure*}[t]
\begin{center}
\includegraphics[width=1.0\linewidth]{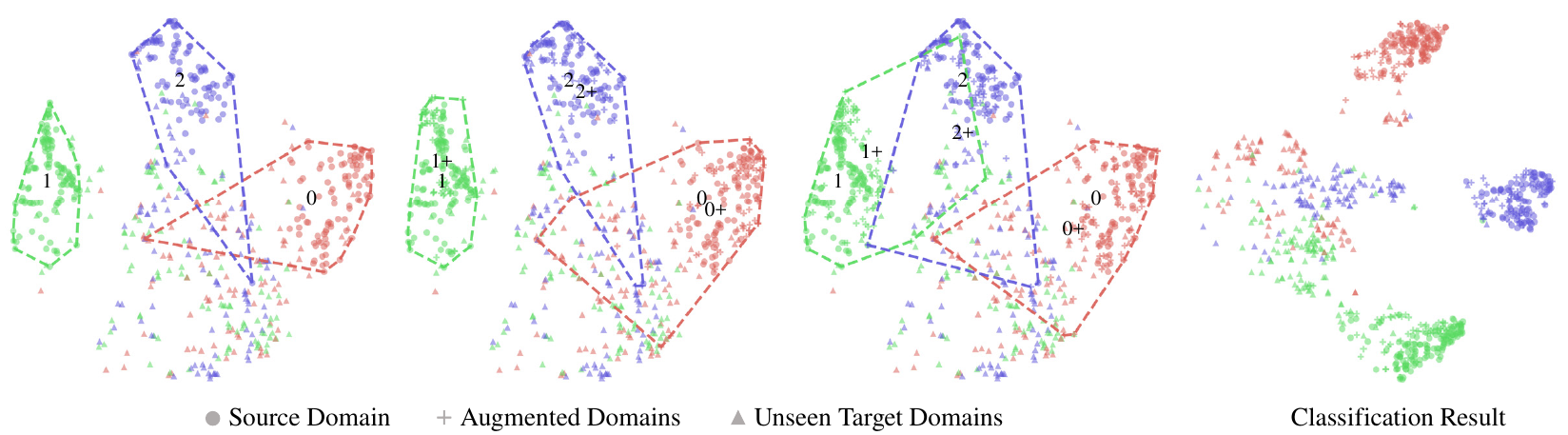}
\end{center}
\vspace{-1em}
\caption{Visualization of domains and convex hulls in the embedding space (the first three figures) and classification space (the last figure).  {\bf From left to right}: (a) source domain $\mathcal{S}$ and unseen target domains $\mathcal{T}$; (b) augmented domains $\mathcal{S}^+$ w/o $\mathcal{L}_{\mathrm{relax}}$; (c) $\mathcal{S}^+$ w/ $\mathcal{L}_{\mathrm{relax}}$; (d) the classification result of SDG. Different colors denote different categories. The numbers mark the corresponding cluster centers. Note that \textbf{1}: cluster center of $\mathcal{S}$; \textbf{1+}: cluster center of $\mathcal{S}^+$. Best viewed in color and zoom in for details.}
\label{fig_tsne}
\end{figure*}

For deep neural networks, Eq.~\eqref{eq_problem} is intractable with arbitrary $\rho$ and $\eta$. Consequently, we consider its Lagrangian relaxation with fixed penalty parameters $\alpha \ge 0$ and $\beta \ge 0$:
\begin{equation}\label{eq:lagrangian-duality}\nonumber
\min_{\theta}\{
\sup_{\mathcal{S}^+} \left \{ \mathbb{E}[\mathcal{L}_{\mathrm{task}}(\theta;\mathbf{x}^+) ] - W_{c,d}\right\} = \mathbb{E}[\phi_{\alpha,\beta}(\theta,\psi;\mathbf{x})]\},
\end{equation}
and we have $W_{c,d}(\mathcal{S}, \mathcal{S}^+) = \alpha W_c(\mathcal{S}, \mathcal{S}^+) -\beta W_d(\mathcal{S}, \mathcal{S}^+)$, 
$\phi_{\alpha,\beta}(\theta,\psi;\mathbf{x})= \sup_{\mathbf{x}^+}\left\{\mathcal{L}_{\mathrm{task}}(\theta;\mathbf{x}^+) -\mathcal{L}_{c,d}  \right\}$, 
and $\mathcal{L}_{c,d} = \alpha c\left(\mathbf{z}, \mathbf{z}^+\right)-\beta d\left(\mathbf{x}, \mathbf{x}^+\right)$.
Thus the problem in Eq.~\eqref{eq_problem} is transformed to minimize the robust surrogate $\phi_{\alpha,\beta}$. 

According to \cite{sinha2017certifying}, $\phi_{\alpha}$ is smooth w.r.t $\theta$ if $\alpha$ is large enough and the assumption of Lipschitzian smoothness holds. Since $\psi$ and $\theta$ are independent with each other, $\phi_{\alpha,\beta}$ is still smooth w.r.t $\theta$. The gradient can be computed as:
\begin{equation}\nonumber
  \nabla_\theta \phi_{\alpha,\beta}(\theta,\psi; \mathbf{x})
  = \nabla_\theta \mathcal{L}_{\mathrm{task}}(\theta; \mathbf{x}^{\star}(\mathbf{x}, \theta, \psi)),
\end{equation}
where $\mathbf{x}^\star(\mathbf{x},\theta,\psi) = \argmax_{\mathbf{x}^+} [\mathcal{L}_{\mathrm{task}}(\theta; \mathbf{x}^+)-\mathcal{L}_{c,d}] = \argmax_{\mathbf{x}^+}\mathcal{L}_\text{ADA}(\theta,\psi;\mathbf{x}^+,\mathbf{z}^+) $, which is exactly the adversarial perturbation defined in Eq.~\eqref{eq:uada}.
\section{Implementation Details}

{\bf Task models:} We design specific task models and employ different training strategies for the three datasets according to their characteristics.

In Digits dataset, the model architecture is \textit{conv-pool-conv-pool-fc-fc-softmax}.
There are two 5 $\times$ 5 convolutional layers with 64 and 128 channels respectively. Each convolutional layer is followed by a max pooling layer with the size of 2 $\times$ 2. The size of the two Fully-Connected (FC) layers is 1024 and the size of the softmax layer is 10.

In CIFAR-10-C~\cite{hendrycks2019benchmarking}, we use Wide Residual Network (WRN)~\cite{zagoruyko2016wide} with 16 layers and the width is 4. The first
layer is a 3$\times$3 convolutional layer. It converts the original image with 3 channels to feature maps of 16 channels. Then the
features go through three groups of 3$\times$3 convolutional layers. Each group consists of two blocks and each block is composed of two convolutional layers with the same number of channels. And their channels are \{64, 128, 256\} respectively. Each convolutional layer is followed by batch normalization (BN)~\cite{ioffe2015batch}. 
An average pooling layer with the size of 8 $\times$ 8 is appended to the output of the third group. Finally, a softmax layer with the size of 10 predicts the distribution over classes.

In SYTHIA~\cite{ros2016synthia}, we use FCN-32s~\cite{long2015fully} with the backbone of ResNet-50~\cite{he2016deep}. The model begins with ResNet-50. 
1$\times$1 convolutional layer with 14 channels is appended to predict scores for each class at each of the coarse output locations. A deconvolution layer is followed to up-sample the coarse outputs to the original size through bilinear interpolation.

{\bf Wasserstein Auto-Encodes:} We follow~\cite{tolstikhin2018wasserstein} to implement WAEs but slightly modifying architectures for the three datasets according to their characteristics.

In Digits dataset, the encoder and decoder are built with FC layers. The encoder consists of two FC layers with the size of 400 and 20 respectively. Accordingly, the decoder consists of two FC layers with the size of 400 and 3072 respectively.
The discriminator consists of two FC layers with the size of 128 and 1 respectively. 
The architecture of is shown in Fig.~\ref{fig_waes}~(a).

In CIFAR-10-C~\cite{hendrycks2019benchmarking}, the encoder begins with four convolutional layers with the channels of \{16, 32, 32, 32\}. And two FC layers with the size of 1024 and 512 are followed. Accordingly, the decoder begins with two FC layers with the size of 512 and 1024 respectively. And four deconvolution layers with the channels of \{32, 32, 16, 3\} are followed.
Each layer is followed by BN~\cite{ioffe2015batch} except for the final layer of the decoder.
The discriminator consists of two FC layers with the size of 128 and 1 respectively. The architecture is shown in Fig.~\ref{fig_waes}~(b).

In SYTHIA~\cite{ros2016synthia}, the encoder begins with three convolutional layers with the channels of \{32, 64, 128\}. And two FC layers with the size of \{3840, 512\} are followed. Accordingly, the decoder begins with two FC layers with the size of \{512, 3840\}. And three deconvolution layers with the channels of \{64, 32, 3\} are followed. Each layer is followed by BN~\cite{ioffe2015batch} except for the final layer of the decoder. 
The discriminator consists of three FC layers with the size of \{512, 512, 1\}. The architecture is shown in Fig.~\ref{fig_waes}~(c).

We apply the Adam optimizer in training WAEs. The learning rate is 0.001 for Digits and 0.0001 for both CIFAR-10-C and SYTHIA.
The training epoches is 20 for Digits, 100 for CIFAR-10-C~\cite{hendrycks2019benchmarking}, and 200 for SYTHIA~\cite{ros2016synthia}.

\section{Experiments}

We begin by introducing the experimental setups in Sec.~\ref{sec:setting}. In Sec.~\ref{sec:ablation}, we carry out detailed ablation study to validate the strength of the proposed relaxation, the efficiency of meta-learning scheme, the selection and trade-off of key hyperparameters, and the proposed uncertainty quantification. In Sec.~\ref{sec:sota}, we compare our method with state of the arts on benchmark datasets.
In Sec.~\ref{sec:few}, we further evaluate our method in {\it few-shot domain adaptation}.

\subsection{Datasets and Settings}\label{sec:setting}

\indent{\bf Datasets and settings:} (1) {\it Digits} consists of five sub-datasets: MNIST \cite{lecun1998gradient}, MNIST-M \cite{ganin2015unsupervised}, SVHN \cite{netzer2011reading}, SYN \cite{ganin2015unsupervised}, and USPS \cite{denker1989advances}, and each of them can be viewed as a different domain. Each image in these datasets contains one single digit with different styles. This dataset is mainly employed for ablation studies.
We use the first 10,000 samples in the training set of MNIST for training, and evaluate models on all other domains.
(2) {\it CIFAR-10-C} \cite{hendrycks2019benchmarking} is a robustness benchmark consisting of 19 corruptions types with five levels of severities applied to the test set of CIFAR-10. The corruptions come from four main categories: noise, blur, weather and digital. Each corruption has five-level severities and ``5'' indicates the most corrupted one. All the models are trained on CIFAR-10 and evaluated on CIFAR-10-C.
(3) {\it SYTHIA} \cite{ros2016synthia} is a dataset synthesized for semantic segmentation in the context of driving scenarios. This dataset consists of the same traffic situation but under different locations (Highway, New York-like City and Old European Town are selected)  and different weather/illumination/season conditions (Dawn, Fog, Night, Spring and Winter are selected). Following the protocol in \cite{volpi2018generalizing}, we only use the images from the left front camera and 900 images are randomly sample from each source domain.

{\bf Evaluation metrics:} For Digits and CIFAR-10-C, we compute the mean 
accuracy on each unseen domain. For CIFAR-10-C, accuracy may not be sufficient to comprehensively evaluate the performance of models without measuring relative gain over baseline models (ERM \cite{koltchinskii2011oracle}) and relative error evaluated on the \textit{clean} dataset, i.e., the test set of CIFAR-10 without any corruption.
Inspired by the robustness metrics proposed in \cite{hendrycks2019benchmarking}, two metrics are formulated to evaluate the robustness against image corruptions in the context of domain generalization: mean Corruption Error (mCE) and Relative mCE (RmCE). They are defined as:
$\mathrm{mCE}=\frac{1}{N}\sum_{i=1}^{N} E_{i}^{f} /E_{i}^{\mathrm{ERM}}$,
$\mathrm{RmCE}=\frac{1}{N} \sum_{i=1}^{N} (E_{i}^{f}-E_{\text {clean }}^{f})/(E_{i}^{\mathrm{ERM}}-E_{\text {clean }}^{\mathrm{ERM}})$,
where $N$ is the number of corruptions. mCE is used for evaluating the robustness of the classifier $f$ compared with ERM \cite{koltchinskii2011oracle}. RmCE measures the relative robustness compared with the \textit{clean} data. For SYTHIA, we compute the standard mean Intersection Over Union (mIoU) on each unseen domain.

\begin{figure*}[t]
\begin{center}
\subfigure{
\includegraphics[width=0.23\linewidth]{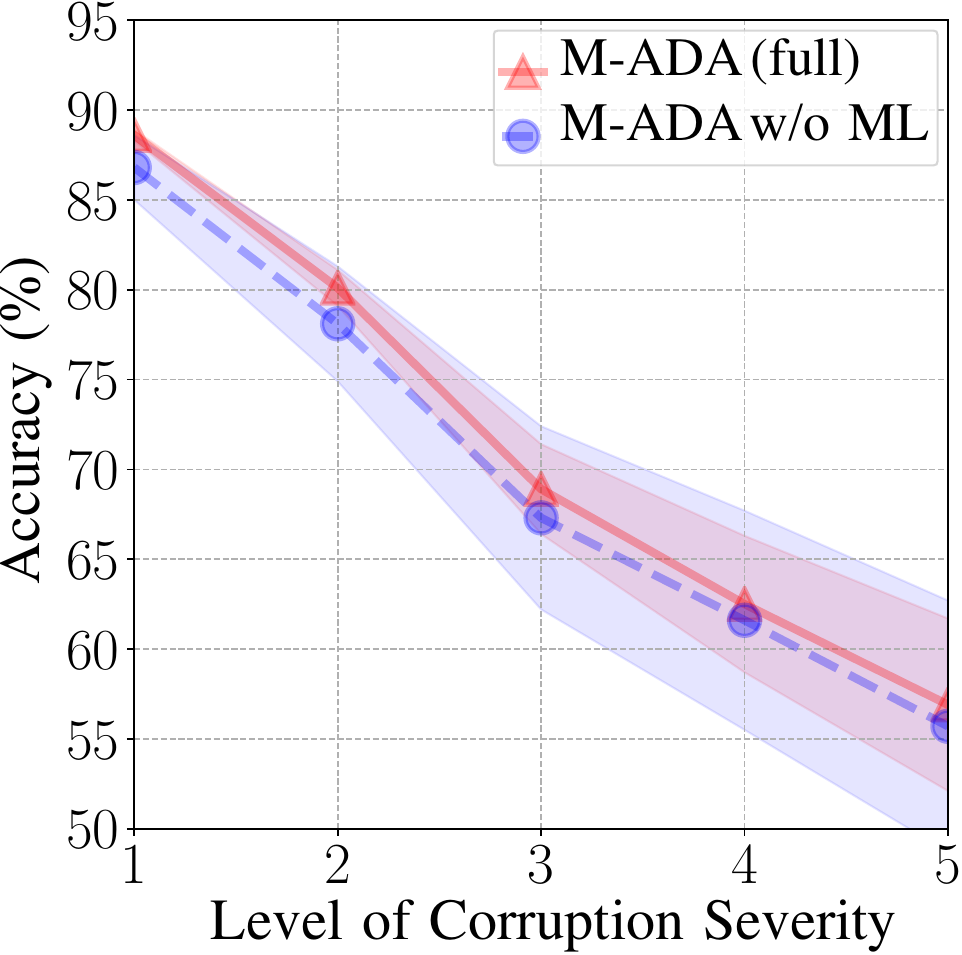}
}
\subfigure{
\includegraphics[width=0.23\linewidth]{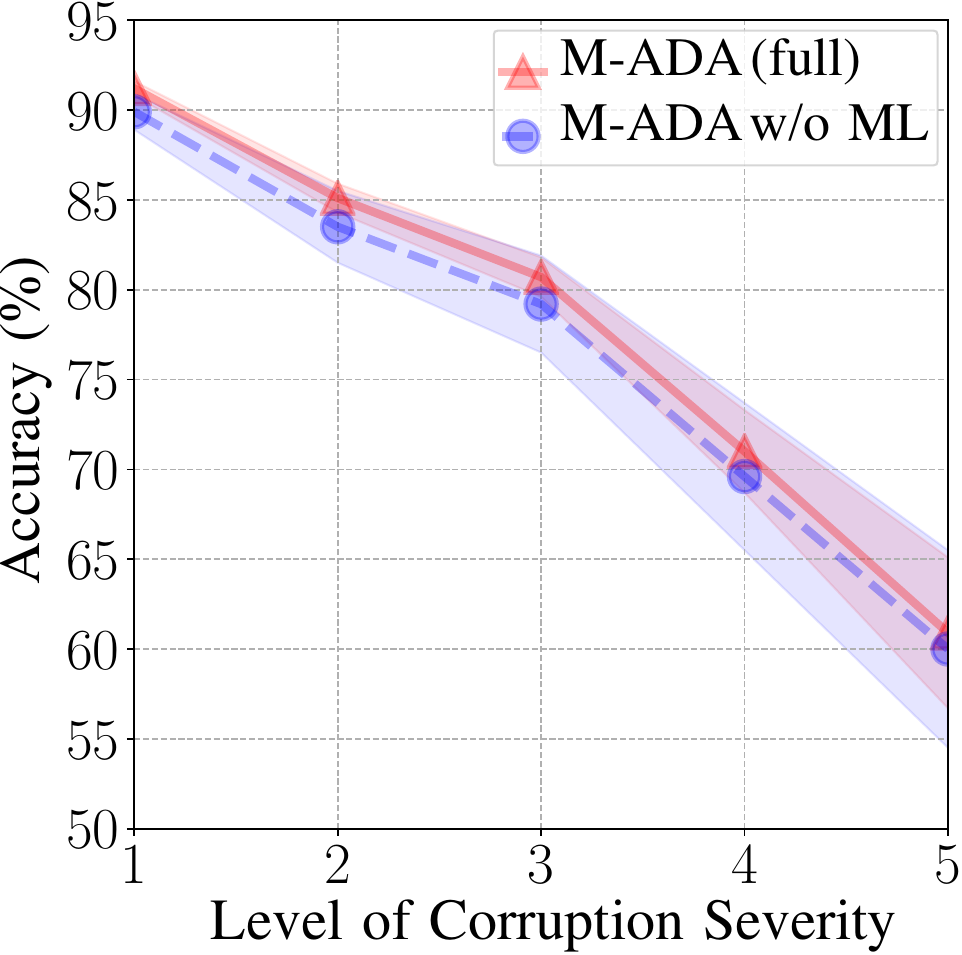}
}
\subfigure{
\includegraphics[width=0.23\linewidth]{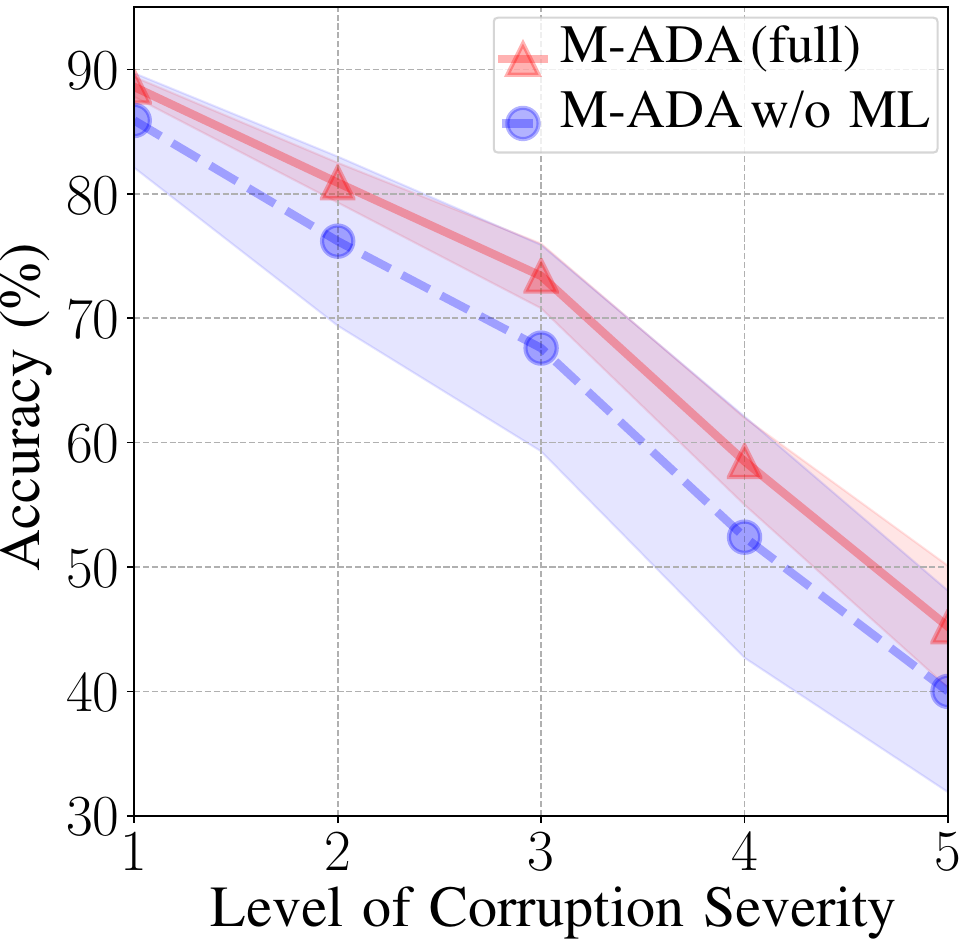}
}
\subfigure{
\includegraphics[width=0.23\linewidth]{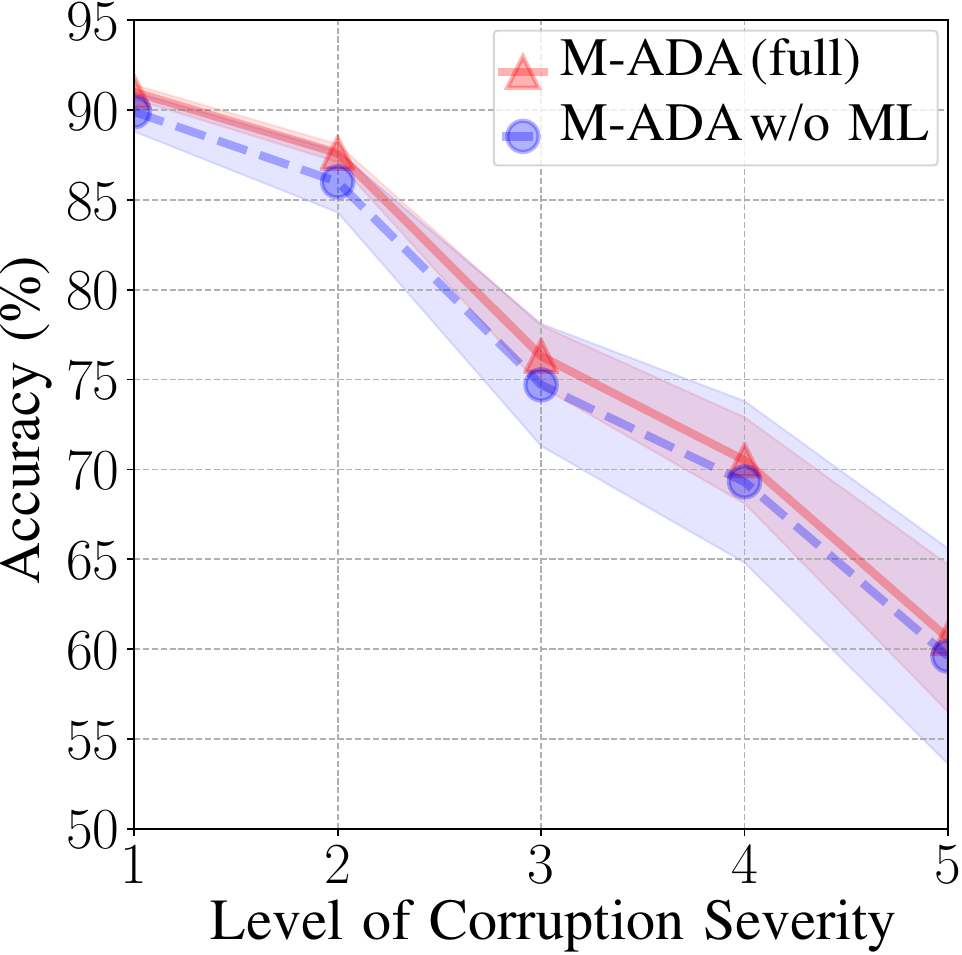}
}
\end{center}
\vspace{-1em}
\caption{Validation of meta-learning scheme. 
Five levels of severity are evaluated on each unseen corruption.
\textbf{From left to right:} (a) \textit{Gaussian Noise}; (b) \textit{Speckle Noise} ; (c) \textit{Impulse Noise}; and (d) \textit{Shot Noise}.
}
\label{fig_cifar10meta}
\end{figure*}

\subsection{Ablation Study}\label{sec:ablation}

In this section, we conduct experiments to evaluate the effect of the proposed relaxation term $\mathcal{L}_{\mathrm{relax}}$ in Eq.~\eqref{eq:uada}, the meta-learning training scheme (Sec.~\ref{sec:meta}), three key hyper-parameters ($K$, $\alpha$, and $\beta$), and the proposed uncertainty quantification (Sec.~\ref{sec:uncertain}).

{\bf Validation of $\mathcal{L}_{\mathrm{relax}}$:} To give an intuitive understanding of how $\mathcal{L}_{\mathrm{relax}}$ affects the distribution of augmented domains $\mathcal{S}^+$, we use t-SNE \cite{maaten2008tsne} to visualize $\mathcal{S}^+$ with and without $\mathcal{L}_{\mathrm{relax}}$ in the embedding space. Their results are shown in Fig.~\ref{fig_tsne}~(b) and~(c), respectively. We observe that the convex hull of $\mathcal{S} \cup$ $\mathcal{S}^+$ with $\mathcal{L}_{\mathrm{relax}}$ covers an enlarged region than that of $\mathcal{S} \cup \mathcal{S}^+$ without $\mathcal{L}_{\mathrm{relax}}$. This indicates that $\mathcal{S}^+$ contains more distributional variance and better overlaps with unseen domains. Further, we compute Wasserstein distance to quantitatively measure the difference between $\mathcal{S}$ and $\mathcal{S}^+$. The distance between $\mathcal{S}$ and $\mathcal{S}^+$ with $\mathcal{L}_{\mathrm{relax}}$ is 0.078, while if $\mathcal{L}_{\mathrm{relax}}$ is not employed, the distance decreases to 0.032, indicating an improvement of $58.9\%$ by introducing $\mathcal{L}_{\mathrm{relax}}$. These results demonstrate that $\mathcal{L}_{\mathrm{relax}}$ is capable of pushing $\mathcal{S}^+$ away from $\mathcal{S}$, which guarantees significant domain transportation in the input space.

\begin{figure*}[t]
\begin{center}
\subfigure{
	\includegraphics[width=0.3\linewidth]{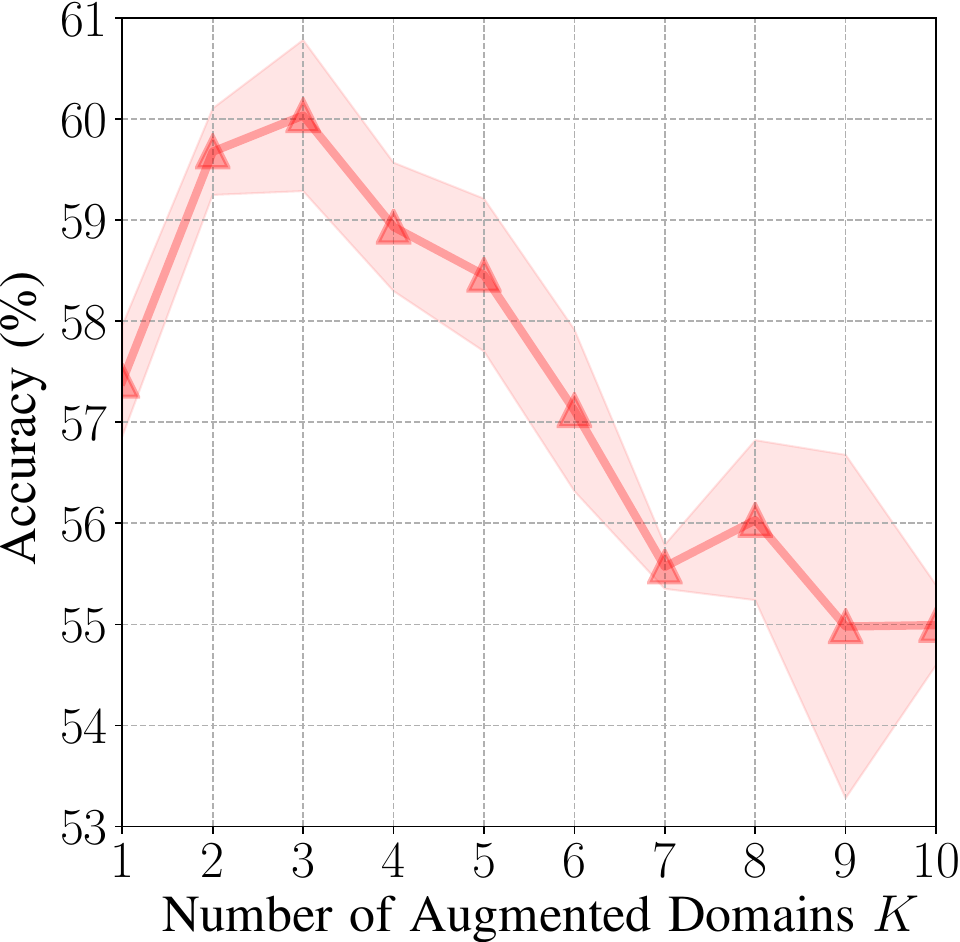}
}
\subfigure{
	\includegraphics[width=0.3\linewidth]{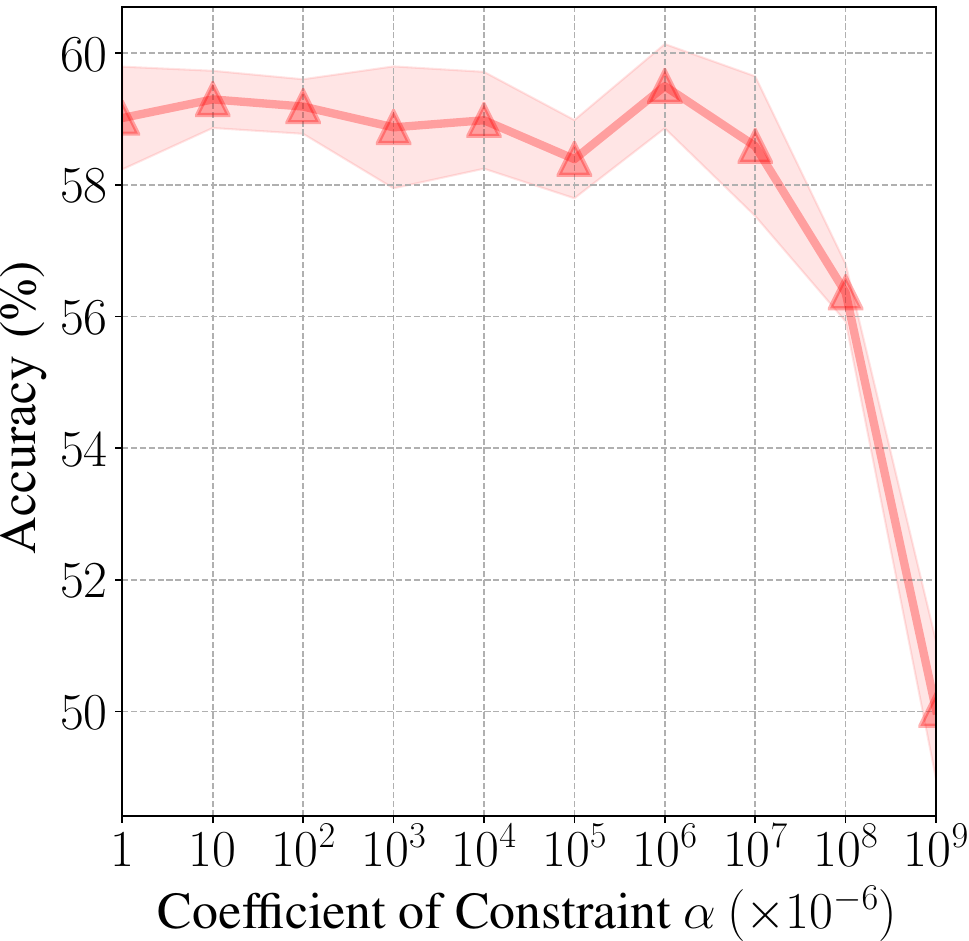}
}
\subfigure{
	\includegraphics[width=0.3\linewidth]{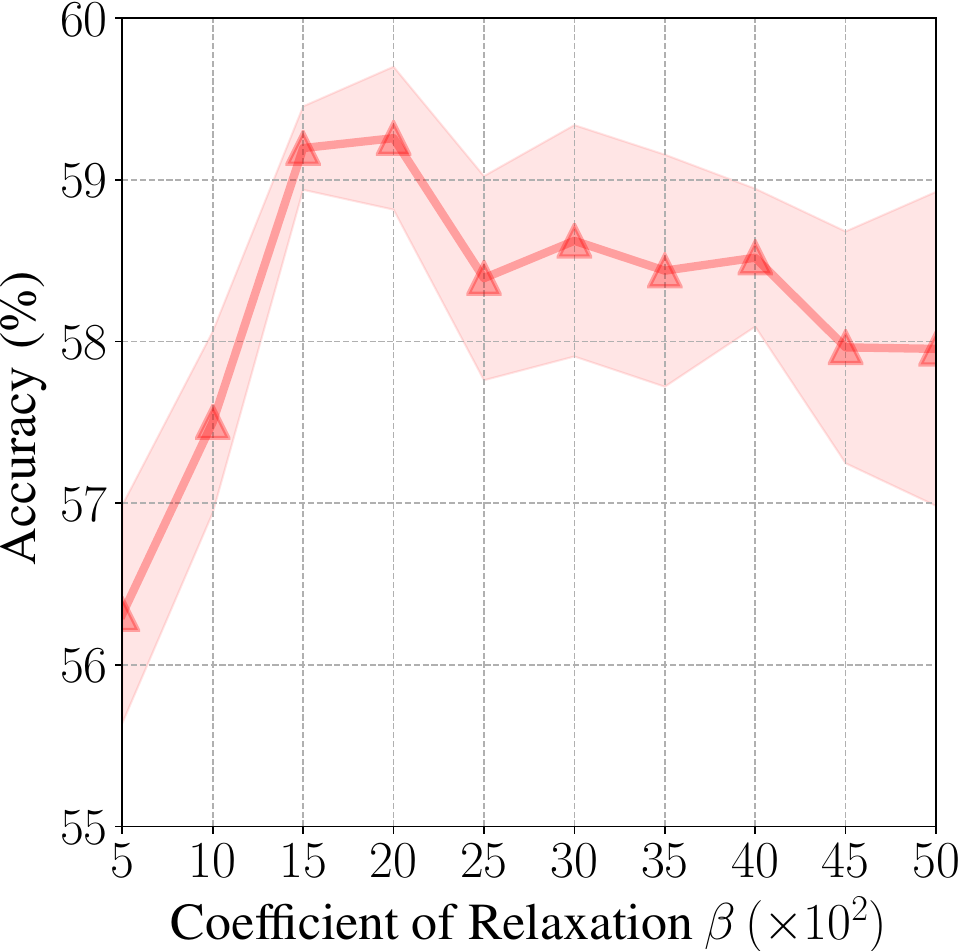}
}
\end{center}
\vspace{-1em}
\caption{Hyper-parameter tuning of $K$, $\alpha$, and $\beta$. We set $K=3$, $\alpha=1.0$, and $\beta=2.0\times10^3$ according to the best accuracy.} 
\label{fig_params}
\end{figure*}

{\bf Validation of meta-learning scheme:} The comparisons of our method with and without meta-learning (ML) scheme are presented in Tabs.~\ref{tab_digits} and~\ref{tab:cifar10_s5}. We observe that with the help of this meta-learning scheme, the results on average accuracy of Digits and CIFAR-10-C are improved by 0.94\% and 1.37\%, respectively. Specially, the results of two kinds of unseen corruptions are shown in Fig.~\ref{fig_cifar10meta}. As seen, the meta-learning scheme can significantly reduce variance and yield better performance across all levels of severity. The experimental results prove that the meta-learning scheme plays a key role to improve the training stability and classification accuracy. This is extremely important when performing adversarial domain augmentation in challenging conditions.

\begin{figure*}
\centering
\begin{minipage}[h]{0.65\textwidth}
\centering
\includegraphics[width=1.\linewidth]{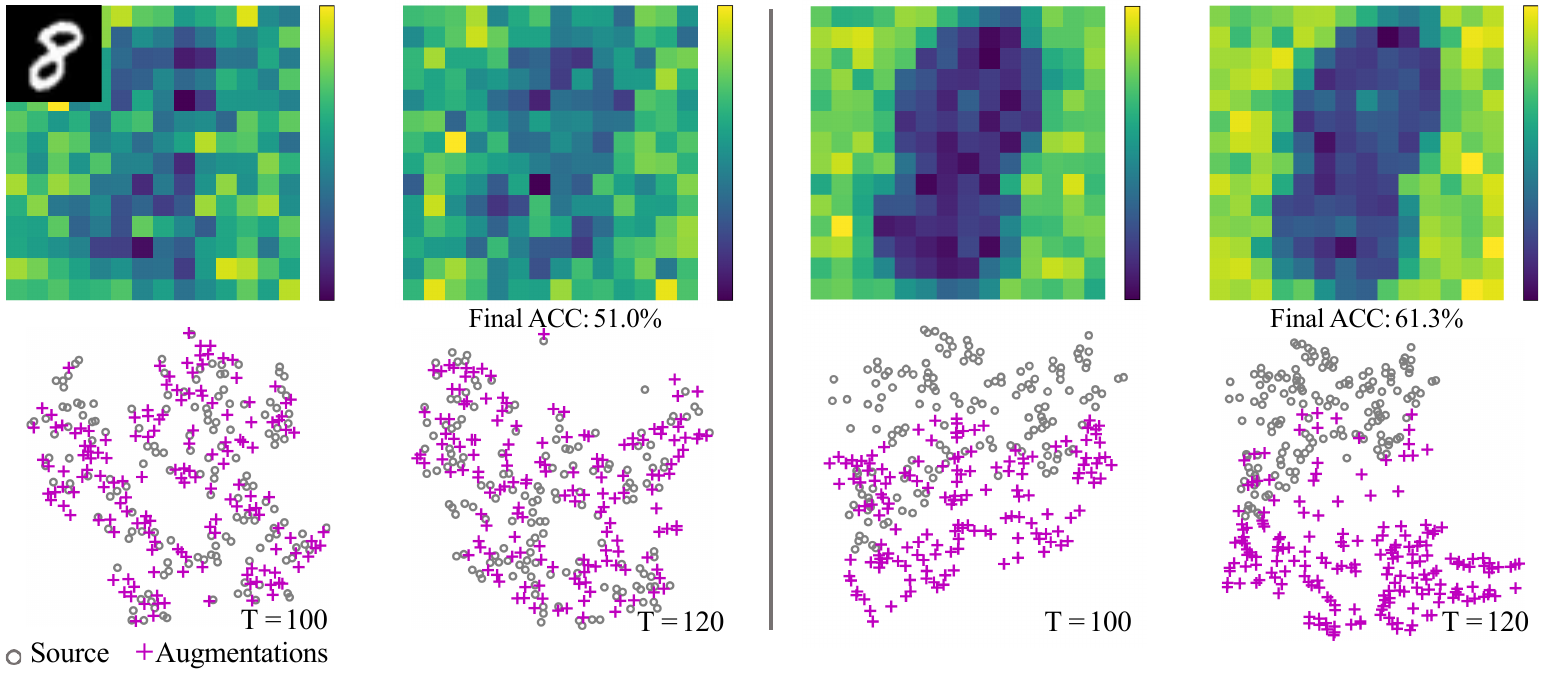}
\caption{Visualization of feature perturbation $|\mathbf{e}| = |\mathbf{h}^+ -\mathbf{h}|$ (\textbf{Top}) and embedding of domains (\textbf{Bottom}) at different training iterations $T$ on \textit{MNIST}.
\textbf{Left:} Models w/o uncertainty; \textbf{Right:} Models w/ uncertainty. Most perturbations are located in the background area and models w/ uncertainty can create large domain transportation in a curriculum learning scheme.}\label{fig:feature}
\end{minipage}
\hspace{0.1in}
\begin{minipage}{0.32\textwidth}
\centering
\includegraphics[width=1.\linewidth]{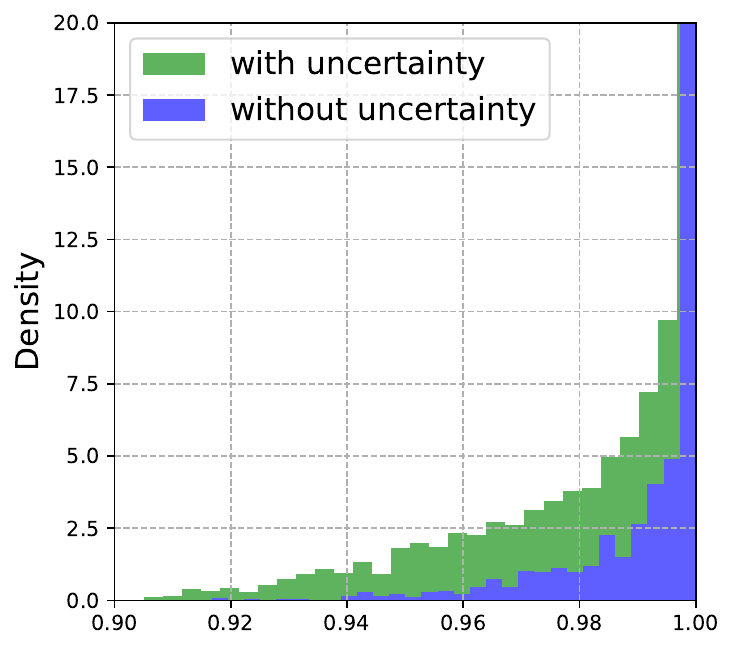}
\caption{Visualization of label mixup $\mathbf{y}^+$ on \textit{MNIST}. Models w/ uncertainty can encourage more smoothing labels and significantly increase the capacity of label space.}\label{fig:y_plus}
\end{minipage}
\end{figure*}

{\bf Hyper-parameter tuning of $K$, $\alpha$, and $\beta$:} We study the effect of three important hyper-parameters: the number of augmented domains ($K$), the distance between the source and augmented domain in the embedding space ($\alpha$), and the deviation between the source and augmented domain ($\beta$). We plot the accuracy curve under different $K$, $\alpha$, and $\beta$ in Fig.~\ref{fig_params}. In Fig.~\ref{fig_params}~(left), we find that the accuracy reaches the summit when $K=3$ and keeps falling with $K$ increasing. This is due to the fact that excessive adversarial samples above a certain threshold will increase the instability and degrade the robustness of the model. Since the distance between the augmented and source domain increases as $K$ increases, a large $K$ may break down the constraint of semantic consistency yielding inferior model training.
In Fig.~\ref{fig_params}~(middle), we find that the accuracy reaches the summit when $\alpha=1.0$ and keeps falling with $\alpha$ increasing. This is because large $\alpha$ will make the source and augmented domain too close in the embedding space, yielding limited domain transportation.
In Fig.~\ref{fig_params}~(right), we observe that the accuracy reaches the summit when $\beta=2.0 \times 10^3$ and drops slightly when $\beta$ increases. This is because large $\beta$ will produce domains too far way from the source $\mathcal{S}$ and even reach out of the manifold in embedding space.

{\bf Validation of uncertainty quantification.}
We visualize feature perturbation $|\mathbf{e}|= |\mathbf{h}^+ -\mathbf{h}|$ and the embedding of domains at different training iterations $T$ on MNIST~\cite{lecun1998gradient}. We use t-SNE~\cite{maaten2008tsne} to visualize the source and augmented domains without and with uncertainty assessment in the embedding space.
Results are shown in Fig.~\ref{fig:feature}.
In the model  without  uncertainty (left), the feature perturbation $\mathbf{e}$ is sampled from $\mathcal{N}(\mathbf{0},\mathbf{I})$ without learnable parameters. In the model  with uncertainty (right),
we observe that most perturbations are located in the background area which increases the variation of $\mathcal{S}^+$ while keeping the category unchanged.
As a result, models with uncertainty can create large domain transportation in a curriculum learning scheme, yielding safe augmentation and improved accuracy on unseen domains.
We visualize the density of $\mathbf{y}^+$ in Fig.~\ref{fig:y_plus}. As seen, models with uncertainty can  significantly augment the label space.

\subsection{Evaluation of Single Domain Generalization}\label{sec:sota}

We compare our method with the following five state-of-the-art methods. 
(1) Empirical Risk Minimization (ERM) \cite{vapnik1998statistical,koltchinskii2011oracle} are models trained with cross-entropy loss, without any auxiliary loss and data augmentation scheme.
(2) CCSA~\cite{motiian2017unified} uses semantic alignment to regularize the learned feature subspace for domain generalization. 
(3) d-SNE~\cite{xu2019dsne} minimizes the largest distance between the samples from the same class and maximizes the smallest distance between the samples from different classes.
(4) GUD~\cite{volpi2018generalizing} proposes an adversarial data augmentation method for single domain generalization, which is the related work to our method.
(5) JiGen~\cite{carlucci2019jigasaw} learns to classify and predict the order of shuffled image patches at the same time for domain generalization.

{\bf Comparison on Digits:} We train all models on MNIST and test them on unseen domains, i.e., MNIST-M, SVHN, SYN, and USPS. We report the results in Tab.~\ref{tab_digits}. We observe that our method outperforms GUD with a large margin on SVHN, MNIST-M and SYN. The improvement on USPS is not as significant as those on other domains, mainly due to its great similarity with MNIST. On the contrary, CCSA and d-SNE obtain large improvements on USPS but perform poorly on other ones. 
Uncertain SDG outperforms SDG on {\it SYN} and the average accuracy by 8.1\% and 1.8\%, respectively.
We also compare Uncertain SDG to SDG in terms of model size and training time. Results on {\it Digits} are shown in Tab.~\ref{tab:efficiency}. 
As observed, Uncertain SDG can reduce $\sim25\%$  parameters and $\sim30\%$ training time. The strong results testify the efficiency of the proposed uncertain single domain generalization.

\begin{table}[t]
\begin{center}
\resizebox{1.\linewidth}{!}{
\begin{tabular}{@{}lccccc@{}}
\toprule
Method               & SVHN  & MNIST-M & SYN & USPS & Avg.  \\
\hline
ERM \cite{koltchinskii2011oracle}                  &  27.83       & 52.72 & 39.65 & 76.94 & 49.29\\
CCSA \cite{motiian2017unified} & 25.89 & 49.29 & 37.31  & 83.72   & 49.05   \\
d-SNE \cite{xu2019dsne} &26.22      &  50.98  & 37.83    &\textbf{93.16 }& 52.05 \\
JiGen  \cite{carlucci2019jigasaw} & 33.80     &  57.80   &43.79  & 77.15   & 53.14 \\
GUD \cite{volpi2018generalizing} & 35.51   &  60.41 & 45.32 & 77.26 &  54.62\\
\hline
SDG w/o $\mathcal{L}_{\mathrm{relax}}$ & 37.33 & 61.43   & 45.58 & 77.37 & 55.43   \\
SDG w/o $\mathcal{L}_{\mathrm{const}}$ & 41.36  & 67.28  & 47.94 & 78.22 &  58.70 \\
SDG w/o ML & 41.45  & \underline{67.86} & 48.76 & 76.12 &  58.55\\
{\bf SDG} (full) &\underline{42.55} & \textbf{67.94} & \underline{48.95} &\underline{78.53} & \underline{59.49}\\
{\bf Uncertain SDG} &\textbf{43.32} & 67.38 & \textbf{57.14} &77.39 & \textbf{61.31}\\
\bottomrule
\end{tabular}}
\end{center}
\caption{Single domain generalization comparison (\%) on {\it Digits}. Models are trained on MNIST. The variant (w/o $\mathcal{L}_{\mathrm{relax}}$) has the most significant performance decrease, indicating it is crucial to perform Wasserstein relaxation for single domain generalization.}\label{tab_digits}
\end{table}
\begin{table}[t]
	\begin{center}
		\resizebox{.85\linewidth}{!}{
		\begin{tabular}{@{}lccccc@{}}
			\toprule
			Method&  \# of params. & Training time  & Accuracy\\
			\hline
			SDG  & 7.02M & 23.7min & 59.5\% \\
			Uncertain SDG & 5.27M &  16.4min & 61.3\% \\
			\bottomrule
	\end{tabular}}
	\caption{ Efficiency comparison in terms of model size and training time. Uncertain SDG outperforms SDG marginally in terms of memory, speed, and accuracy.}\label{tab:efficiency}
	\end{center}
\end{table}
\begin{table}[t]
	\begin{center}
		\resizebox{1.\linewidth}{!}{
		\begin{tabular}{@{}lccccc@{}}
			\toprule
			Method& Level 1& Level 2& Level 3& Level 4& Level 5 \\
			\hline
			ERM \cite{koltchinskii2011oracle}& 87.8$\pm$0.1& 81.5$\pm$0.2& 75.5$\pm$0.4& 68.2$\pm$0.6& 56.1$\pm$0.8 \\
			GUD \cite{volpi2018generalizing}& 88.3$\pm$0.6& 83.5$\pm$2.0& 77.6$\pm$2.2& 70.6$\pm$2.3& 58.3$\pm$2.5 \\
			\hline
			{\bf SDG} (full) &\textbf{90.5$\pm$0.3}& \textbf{86.8$\pm$0.4}& \textbf{82.5$\pm$0.6}& \textbf{76.4$\pm$0.9}& \textbf{65.6$\pm$1.2} \\
			$\uparrow$ to ERM &3.08\% & 6.50\%& 9.27\%& 12.0\%& 16.9\% \\
			$\uparrow$ to GUD &2.49\%& 3.95\%& 6.31\%& 8.22\%& 12.5\% \\
			\bottomrule
	\end{tabular}}
	\end{center}
	\caption{Accuracy comparison (\%) on {\it CIFAR-10-C}. Boosts ($\uparrow$) become more significant as corruption severity level ({\bf 1-5}) increases.}\label{tab:cifar10-c_severity}
\end{table}
\begin{table*}[t]
\begin{center}
\resizebox{1.\linewidth}{!}{
\begin{tabular}{@{}lcccccccccccc@{}}
\toprule
&\multicolumn{3}{c}{Weather}& \multicolumn{5}{c}{Blur}& \multicolumn{4}{c}{Noise}\\
\cmidrule(lr){2-4} \cmidrule(lr){5-9} \cmidrule(lr){10-13}
&Fog& Snow& Frost& Zoom& Defocus& Glass& Gaussian& Motion& Speckle& Shot& Impulse& Gaussian\\
\hline
ERM \cite{koltchinskii2011oracle}&65.92& 74.36& 61.57& 59.97& 53.71& 49.44& 30.74& 63.81& 41.31& 35.41& 25.65& 29.01  \\
CCSA \cite{motiian2017unified}&66.94& 74.55& 61.49& 61.96& 56.11& 48.46& 32.22& \textbf{64.73}& 40.12& 33.79& 24.56& 27.85 \\
d-SNE \cite{xu2019dsne}&65.99& 75.46& 62.25& 58.47& 53.71& 50.48& 33.06& 63.70& 45.30& 39.93& 27.95& 34.02 \\
GUD \cite{volpi2018generalizing}&68.29& 76.75& 69.94& 62.95& 56.41& 53.45& 38.33& 63.93& 38.45& 36.87& 22.26& 32.43 \\
\hline
SDG w/o $\mathcal{L}_{\mathrm{relax}}$ &66.99& 80.09& 74.93& 54.15& 44.67& 60.57& 30.53& 57.06& 59.88& 59.18& 43.46& 55.07 \\
SDG w/o ML&67.68& \textbf{80.91}& 76.20& 65.70& 56.87& \textbf{62.14}& 41.20& 63.86& 60.01& 59.63& 40.04& 55.70  \\
{\bf SDG} (full)&\textbf{69.36}& 80.59& \textbf{76.66}& \textbf{68.04}& \textbf{61.18}& 61.59& \textbf{47.34}& 64.23& \textbf{60.88}& \textbf{60.58}& \textbf{45.18}& \textbf{56.88}\\
\bottomrule
\end{tabular}}

\vspace{1em}
\resizebox{0.87\linewidth}{!}{
\begin{tabular}{@{}lcccccccccc@{}}
\toprule
&\multicolumn{7}{c}{Digital}& \\
\cmidrule(lr){2-8}
&Jpeg& Pixelate& Spatter& Elastic& Brightness& Saturate& Contrast& Avg.& mCE &RmCE \\
\hline
ERM \cite{koltchinskii2011oracle}&69.90& 41.07& 75.36& 72.40& \textbf{91.25}& 89.09& \textbf{36.87}& 56.15&  1.00&1.00  \\
CCSA \cite{motiian2017unified}&69.68& 40.94& 77.91& 72.36& 91.00& \textbf{89.42}& 35.83& 56.31&   0.99&0.99 \\
d-SNE \cite{xu2019dsne}&70.20& 38.46& 73.40& 73.33& 90.90& 89.27& 36.28& 56.96&  0.99 &1.00 \\
% JiGen  \cite{carlucci2019jigasaw}&& & & & & & & & & & & & & & \\
GUD \cite{volpi2018generalizing}&74.22& \textbf{53.34}& 80.27& 74.64& 89.91& 82.91& 31.55& 58.26&  0.97&0.95 \\
\hline
SDG w/o $\mathcal{L}_{\mathrm{relax}}$ &76.45& 53.13& 80.75& 73.85& 90.86& 87.01& 27.83& 61.92&  0.90&0.86 \\
SDG w/o ML&\textbf{77.62}& 52.49& \textbf{81.02}& 75.54& 90.69& 86.58& 26.30& 64.22&  0.85 &0.80  \\
{\bf SDG} (full)&77.14& 52.25& 80.62& \textbf{75.61}& 90.78& 87.62& 29.71& \textbf{65.59}&  \textbf{0.82}&\textbf{0.77}\\
\bottomrule
\end{tabular}}
\end{center}
\caption{Robustness comparison on {\it CIFAR-10-C} \cite{hendrycks2019benchmarking}. The models are generalized from clean data to different corruptions. We report the classification accuracy (\%) of 19 corruptions under the corruption level of ``5'' (severest). We also report the mean Corruption Error (mCE) and relative mCE (RmCE) in the last two columns. The lower the better for mCE and RmCE. 
}
\label{tab:cifar10_s5}
\end{table*}
\begin{table*}[th]
\begin{center}
\resizebox{.92\linewidth}{!}{
\begin{tabular}{@{}llccccccccccc@{}}
\toprule
& &\multicolumn{5}{c}{New York-like City}& \multicolumn{5}{c}{Old European Town}& \\
\cmidrule(lr){3-7} \cmidrule(lr){8-12}
Source Domain& Method &Dawn&Fog&Night&Spring&Winter& Dawn&Fog&Night&Spring&Winter& Avg. \\
\hline
\multirow{3}*{Highway/Dawn}&ERM \cite{koltchinskii2011oracle}& 27.80&2.73&0.93&6.80&1.65&52.78&31.37&15.86&33.78&13.35&18.70 \\
&GUD \cite{volpi2018generalizing}& 27.14&4.05&1.63&7.22&2.83&52.80&34.43&18.19&33.58&14.68&19.66 \\
&{\bf SDG}& \underline{29.10}&\underline{4.43}&\textbf{4.75}&\textbf{14.13}&\underline{4.97}&\underline{54.28}&\underline{36.04}&\underline{23.19}&\textbf{37.53}&\underline{14.87}&\underline{22.33} \\
&{\bf Uncertain SDG}& \textbf{29.31} & \textbf{7.62} & \underline{2.84} & \underline{12.70} & \textbf{10.18} & \textbf{54.89} & \textbf{37.03} & \textbf{25.30} & \underline{37.16} & \textbf{17.71} & \textbf{23.47}  \\
\hline
\multirow{3}*{Highway/Fog}&ERM \cite{koltchinskii2011oracle}& 17.24&34.80&12.36&26.38&11.81&33.73&55.03&26.19&41.74&12.32&27.16 \\
&GUD \cite{volpi2018generalizing}& 18.75&\underline{35.58}&\underline{12.77}&26.02&13.05&37.27&\underline{56.69}&28.06&\underline{43.57}&\textbf{13.59}&28.53 \\
&{\bf SDG}& \underline{21.74}&32.00&9.74&\underline{26.40}&\underline{13.28}&\underline{42.79}&56.60&\textbf{31.79}&42.77&12.85&\underline{29.00} \\
&{\bf Uncertain SDG}&\textbf{23.03} &\textbf{36.21}  &\textbf{13.48}  &\textbf{27.64} &\textbf{14.23} &\textbf{43.08} &\textbf{57.44} &\underline{31.04} &\textbf{44.62} &\underline{13.13} &\textbf{ 30.39 } \\
\hline
\multirow{3}*{Highway/Spring}&ERM \cite{koltchinskii2011oracle}& 26.75&26.41&18.22&32.89&24.60&51.72&51.85&35.65&54.00&\underline{28.13}&35.02 \\
&GUD \cite{volpi2018generalizing}& 28.84&29.67&20.85&35.32&27.87&52.21&\textbf{52.87}&35.99&\underline{55.30}&\textbf{29.58}&36.85 \\
&{\bf SDG}& \underline{29.70}&\underline{31.03}&\underline{22.22}&\underline{38.19}&\underline{28.29}&\textbf{53.57}&51.83&\underline{38.98}&\textbf{55.63}&25.29&\underline{37.47} \\
&{\bf Uncertain SDG}& \textbf{30.18}&\textbf{31.75}&\textbf{22.32}&\textbf{38.64}&\textbf{28.81}&\underline{52.98}&\underline{51.96}&\textbf{40.17}&55.17&26.34&\textbf{37.83} \\
\bottomrule
\end{tabular}}
\end{center}
\caption{Semantic segmentation comparison on \textit{SYNTHIA} \cite{ros2016synthia}. The models are generalized from one source domain to many unseen environment settings. We report the standard mean Intersection Over Unions (mIoUs) and demonstrate visual results in Fig.~\ref{fig:segmentation}.}\label{tab:seg}
\end{table*}

{\bf Comparison on CIFAR-10-C:} We train all models on the clean data, {\it i.e.}, CIFAR-10, and test them on the corruption data, {\it i.e., CIFAR-10-C}. In this case, there are totally $19$ unseen testing domains. Results on CIFAR-10-C across five levels of corruption severity are shown in Tab.~\ref{tab:cifar10-c_severity}. As seen, The gap between GUD and our method gets larger with the level of severity increasing, and our method can significantly reduce standard deviations across all levels. In addition, we present the result of each corruption with the highest severity in Tab.~\ref{tab:cifar10_s5}. We observe that our method substantially outperforms other methods on most corruptions. Specially, in several corruptions such as \textit{Snow}, \textit{Glass blur}, \textit{Pixelate} and corruptions related with \textit{Noise}, our method outperforms ERM \cite{koltchinskii2011oracle} with more than 10\%. More importantly, our method has the lowest values on mCE and RmCE, indicating its strong robustness against image corruptions.

\begin{figure*}[t]
\begin{center}
\includegraphics[width=0.95\linewidth]{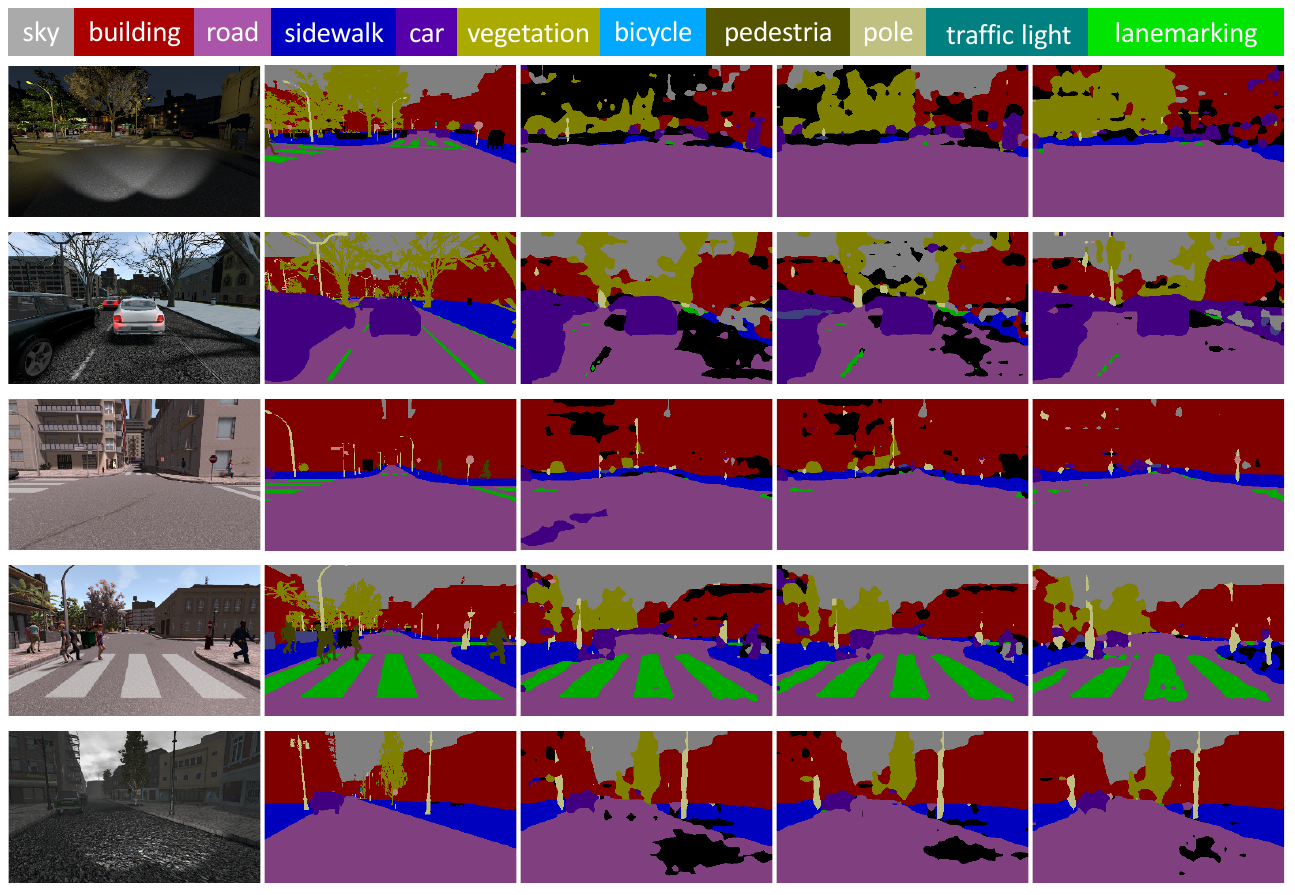}
\end{center}
\caption{Examples of semantic segmentation on {\it SYTHIA}~\cite{ros2016synthia}. {\bf From left to right:} (a) images from unseen domains; (b) ground truth; (c) results of ERM~\cite{koltchinskii2011oracle}; (d) results of SDG; and (e) results of Uncertain SDG. Best viewed in color and zoom in for details.} 
\label{fig:segmentation}
\end{figure*}

{\bf Comparison on SYTHIA:} In this experiment, Highway is the source domain, and New York-like City together with Old European Town are unseen target domains. We report semantic segmentation results in Tab.~\ref{tab:seg} and show some examples in Fig.~\ref{fig:segmentation}. Unseen domains are from different locations and other conditions. We observe that our method outperforms ERM \cite{koltchinskii2011oracle} and GUD \cite{volpi2018generalizing} on average mIoUs across three source domains, suggesting its capability of coping with changes of locations, weather and time.
Improvements over ERM \cite{koltchinskii2011oracle} and GUD \cite{volpi2018generalizing} are not significant compared with the other two datasets, mainly owing to the limited number of training images and high reliance of unseen domains. Uncertain SDG outperforms SDG in most unseen environments.
Results demonstrate that uncertainty quantification can further improve the generalization over unseen domains.

\begin{table}[th]
	\begin{center}
		\resizebox{.94\linewidth}{!}{
		\begin{tabular}{@{}lccccc@{}}
			\toprule
			Method & $\lvert  \mathcal{T} \rvert$  & U \ $\rightarrow$ M & M \ $\rightarrow$ S &  S \ $\rightarrow$ M & Avg.\\
			\hline
			I2I \cite{murez2018image} & \multirow{5}{*}{All}     & 92.20 & - & 92.10 & -\\
			DIRT-T \cite{shu2018dirt} & &   - &54.50 & \textbf{99.40}&-\\
			SE \cite{french2017self}& &    \textbf{98.07} & 13.96 &99.18& 70.40\\
			SBADA \cite{russo2018source} & &   97.60 & \textbf{61.08} &76.14& 78.27\\
			G2A \cite{sankaranarayanan2018generate}&    & 90.80 & - & 92.40& -\\
			\hline
			FADA \cite{motiian2017few} & 7  & 91.50& 47.00 & 87.20&  75.23\\
			CCSA \cite{motiian2017unified} & 10  & 95.71 &37.63 & 94.57 & 75.97\\
			\hline
			\multirow{3}{*}{\bf SDG} & 0  & 71.19 & 36.61 & 60.14 & 55.98\\
			&7 & 92.33 & 56.33  & 89.90  & 79.52\\
			&10 &  93.67  & 57.16 & 91.81 & 80.88\\
			\hline
			\multirow{2}{*}{\bf Uncertain SDG}
		    &7 & 92.97  & 58.12  & 89.30 & 80.13\\
			&10 & 93.16 & 59.77  & 91.67 & \textbf{81.53} \\
			\bottomrule
		\end{tabular}}
	\end{center}
\caption{Few-shot domain adaptation comparison on {\it MNIST(M), USPS(U), and SVHN(S)} in terms of accuracy (\%).  $\lvert \mathcal{T} \rvert$ denotes the number of target samples (per class) used during model training.}\label{tab:few-shot}
\end{table}

\subsection{Evaluation of Few-Shot Domain Adaptation}\label{sec:few}

Although our method is designed for single domain generalization, as mentioned in Sec.~\ref{sec:meta}, we also show that our method can be easily applied for few-shot domain adaptation~\cite{motiian2017few}.

{\bf Settings:} 
In few-shot learning, models are usually first pre-trained on the source domain $\mathcal{S}$ and then fine-tuned on the target domain $\mathcal{T}$. More specifically, we first train our model on $\mathcal{S}$ using all training images. Then we randomly pick out 7 or 10 images per class from $\mathcal{T}$. These images are used to fine-tune the pre-trained model with a learning rate of 0.0001 and a batch size of 16.

{\bf Discussions:} We compare our method with the state-of-the-art methods for few-shot domain adaptation. We also report the results of some unsupervised methods which use images in the target domain for training. Results on MNIST, USPS, and SVHN are shown in Tab.~\ref{tab:few-shot}. We observe that our method obtains competitive results compared with FADA~\cite{motiian2017few} and CCSA~\cite{motiian2017unified}. And our method also outperforms several unsupervised methods which take advantage of unlabeled images from the target domain.
Uncertain SDG achieves the best performance on the average of three tasks. 
The result on the hardest task (\textit{M}$\rightarrow$\textit{S}) is even competitive to that of SBADA \cite{russo2018source}.
Moreover, it is worth noting that both FADA~\cite{motiian2017few} and CCSA~\cite{motiian2017unified} are trained in a manner where samples from $\mathcal{S}$ and $\mathcal{T}$ are strongly coupled. This means that when the target domain changes, an entirely new model has to be trained. On the other hand, for a new target domain, our method only needs to fine-tune the pre-trained model with a few samples within a small number of iterations. This demonstrates the high flexibility of our method.

\section{Conclusion}

In this paper, we present Meta-Learning based Adversarial Domain Augmentation to address the problem of single domain generalization. 
The core idea is to use a meta-learning based scheme for efficiently organizing the training of augmented ``fictitious'' domains, which are OOD from source domain and created by adversarial training.
We further improve our method by integrating uncertainty quantification for broad and safe domain generalization.
In addition to the superior performances we achieved through these experiments, a series of ablation studies further validate the effectiveness of key components in our method. In the future, we expect to extend our work to semi-supervised learning or knowledge transferring in multimodal learning.

\section*{Acknowledgments}\label{sec:acknowledgments}
This work is partially supported by National Science Foundation (NSF) CMMI-2039857 D-ISN-1 and Google Research.

% \appendices

% % you can choose not to have a title for an appendix
% % if you want by leaving the argument blank
% \section{}

% \ifCLASSOPTIONcompsoc
%   % The Computer Society usually uses the plural form
%   \section*{Acknowledgments}
% \else
%   % regular IEEE prefers the singular form
% \fi


\ifCLASSOPTIONcaptionsoff
  \newpage
\fi



%\bibliographystyle{IEEEtran}

% H.~Kopka and P.~W. Daly, \emph{A Guide to {\LaTeX}}, 3rd~ed.\hskip 1em plus

% \end{thebibliography}

{\small
\bibliographystyle{abbrv}
\bibliography{egbib}
}

% \begin{IEEEbiography}[{\includegraphics[width=1in,height=1.25in,clip,keepaspectratio]{mshell}}]{Michael Shell}


%\vfill


\end{document}